\documentclass[runningheads]{llncs}

\usepackage{eccv}

\usepackage{eccvabbrv}

\usepackage{graphicx}
\usepackage{booktabs}
\usepackage{algorithm,algorithmicx,algpseudocode}
\usepackage{caption}

\newcommand{\grad}{\nabla}
\usepackage{multirow}

\usepackage[accsupp]{axessibility}  % Improves PDF readability for those with disabilities.

\usepackage{hyperref}
\usepackage{graphicx}

\usepackage{orcidlink}

\begin{document}

% ---------------------------------------------------------------
% TODO REVIEW: Replace with your title
\title{LogicalDefender: Discovering, Extracting, and Utilizing Common-Sense Knowledge} 

% TODO REVIEW: If the paper title is too long for the running head, you can set
% an abbreviated paper title here. If not, comment out.
\titlerunning{LogicalDefender}

% TODO FINAL: Replace with your author list. 
% Include the authors' OCRID for the camera-ready version, if at all possible.
\author{Yuhe Liu\inst{1,2}\thanks{Work done during an internship at Meituan.} \quad\quad
Mengxue Kang\inst{2} \quad\quad
Zengchang Qin\inst{1} \quad\quad
Xiangxiang Chu\inst{2}}

% TODO FINAL: Replace with an abbreviated list of authors.
\authorrunning{Liu et al.}
% First names are abbreviated in the running head.
% If there are more than two authors, 'et al.' is used.

% TODO FINAL: Replace with your institution list.
\institute{Beihang University \and
Meituan Inc.}

\maketitle

\vspace{-1cm}
\begin{figure}[htbp]
\centering
% 第一行
\rotatebox{90}{initial} % 行标注
\begin{minipage}{.21\linewidth}
  \includegraphics[width=\linewidth]{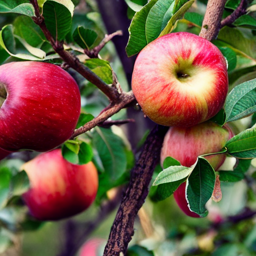}
\end{minipage}%
\hspace{1mm}
\begin{minipage}{.21\linewidth}
  \includegraphics[width=\linewidth]{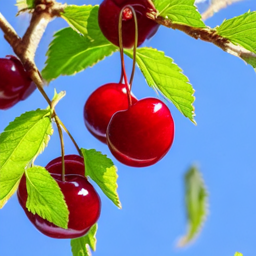}
\end{minipage}%
\hspace{1mm}
\begin{minipage}{.21\linewidth}
  \includegraphics[width=\linewidth]{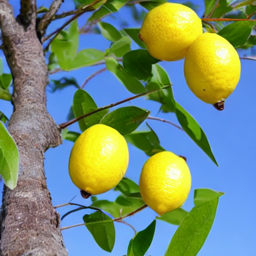}
\end{minipage}%
\hspace{1mm}
\begin{minipage}{.21\linewidth}
  \includegraphics[width=\linewidth]{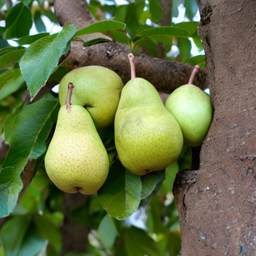}
\end{minipage}
\vspace{1mm}

% 第二行
\rotatebox{90}{rules} % 行标注
\begin{minipage}{.21\linewidth}
  \includegraphics[width=\linewidth]{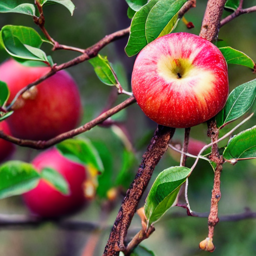}
\end{minipage}%
\hspace{1mm}
\begin{minipage}{.21\linewidth}
  \includegraphics[width=\linewidth]{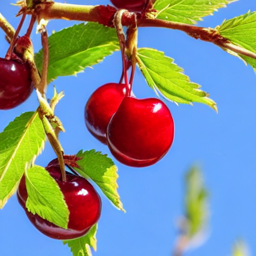}
\end{minipage}%
\hspace{1mm}
\begin{minipage}{.21\linewidth}
  \includegraphics[width=\linewidth]{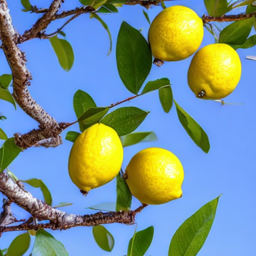}
\end{minipage}%
\hspace{1mm}
\begin{minipage}{.21\linewidth}
  \includegraphics[width=\linewidth]{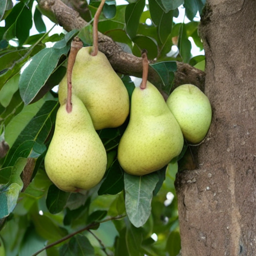}
\end{minipage}
\vspace{1mm}

% 第三行
\rotatebox{90}{ours} % 行标注
\begin{minipage}{.21\linewidth}
  \includegraphics[width=\linewidth]{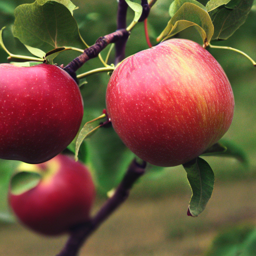}
\end{minipage}%
\hspace{1mm}
\begin{minipage}{.21\linewidth}
  \includegraphics[width=\linewidth]{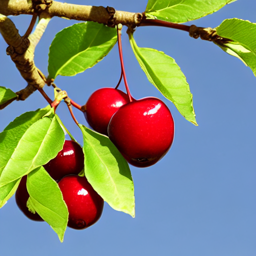}
\end{minipage}%
\hspace{1mm}
\begin{minipage}{.21\linewidth}
  \includegraphics[width=\linewidth]{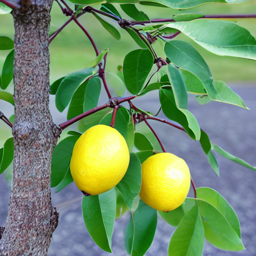}
\end{minipage}%
\hspace{1mm}
\begin{minipage}{.21\linewidth}
  \includegraphics[width=\linewidth]{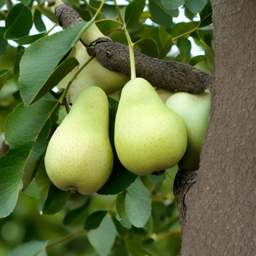}
\end{minipage}

% 列标注
\hspace{1.6cm}apples \hspace{1.6cm}cherries \hspace{1.6cm} lemons\hspace{1.6cm}pears\hspace{1.6cm}
\caption{The images illustrate the logical performance of various types of objects. The first row represents the initial results, referring to direct inference from SD1.5 \cite{LDM}. The second row depicts the rules' results, referring to direct inference from SD1.5 with prompt adjustment. The third row represents results from our method, \emph{LogicalDefender}. The results demonstrate that our method significantly outperforms the other two in terms of logical accuracy. As for apples and pears, the stem in our version is attached to the tree, unlike the first two images where it points outward. As for cherries, our cherries are paired together, attached to the tree by a branch, not simply hanging on the tree as in the first two images. As for lemons, our lemons are connected to the tree via a stem, unlike the first two images where it is suspended in mid-air without connection. }
% As for pears, the stem of our pears is connected to the tree, unlike the first two images where the stem is not attached to anything, pointing aimlessly towards the sky.}
\label{fig:init+final}
\end{figure}

\vspace{-1cm}

\begin{abstract}
Large text-to-image models have achieved astonishing performance in synthesizing diverse and high-quality images guided by texts. With detail-oriented conditioning control, even finer-grained spatial control can be achieved. However, some generated images still appear unreasonable, even with plentiful object features and a harmonious style. In this paper, we delve into the underlying causes and find that deep-level logical information, serving as common-sense knowledge, plays a significant role in understanding and processing images. Nonetheless, almost all models have neglected the importance of logical relations in images, resulting in poor performance in this aspect. Following this observation, we propose \emph{LogicalDefender}, which combines images with the logical knowledge already summarized by humans in text. This encourages models to learn logical knowledge faster and better, and concurrently, extracts the widely applicable logical knowledge from both images and human knowledge. Experiments show that our model has achieved better logical performance, and the extracted logical knowledge can be effectively applied to other scenarios.
\end{abstract}

\section{Introduction}
\label{intro}
Many of us have experienced the ``aha'' moment when we uncover consistent patterns deeply hidden in images and incorporate them into our knowledge base. Consider how we would share the discoveries with others. We would typically distill our findings into language, perhaps supplemented with images for clarity, much like the content of a primary school textbook. (Thus guess why the textbook is designed in this way!) This process actually forms the basis of common-sense knowledge in our society, which has been predominantly transmitted through text over time. In other words, \textbf{language serves as the key tool for carrying condensed knowledge derived from images and everyday experiences, particularly in terms of common-sense and logical principles.}

With the significant advancements in generative models, we can now establish a more direct link between text and images. We can learn complex empirical distributions of images from a collection of billions of image-caption pairs. This enables us to generate images containing specified objects based on a text prompt. 
However, text-to-image models have limitations in logically controlling the spatial composition of the image. Truly understanding and summarizing all consistent logical laws hidden in images can be challenging through training alone. 
Additionally, we find that a substantial portion of the generated images \textbf{do not} adhere to logical or physical laws, as shown in Fig.\ref{fig:init+final} row1.

Our key insight is that the \textbf{logical performance of generated images needs to be taken seriously}, as it is closely related to the basic rules of the real world and indicates models' ability to extract and understand these regulations. However, incorporating these rules into large-scale models often proves challenging. Re-training a model not only requires a large dataset but also involves considerable time consumption. Direct fine-tuning on limited images may lead to overfitting and catastrophic forgetting. Moreover, how to design the model architecture to focus on the logical laws existed in images is challenging.

Given this insight, a natural question arises: \emph{How can we enhance models' logical generation ability under these constraints?} As mentioned earlier, logical laws can be explicitly extracted from common-sense knowledge and displayed as textual explanations. Thus, providing additional text as a prompt during inference could be a straightforward solution. However, this attempt is ineffective. Even though semantic prior is learned from vast image-caption pairs, such semantic prior mainly binds the word with the corresponding features of diverse instances, for example, binding ``apple'' with the appearance of different apples. This approach cannot extract the underlying logical relationships among instances, thus limiting the expressiveness and instruction-following ability (shown in Fig.\ref{fig:init+final} row2 ). In ControlNet\cite{controlnet}, finer-grained control is enabled by providing images which directly specify the spatial composition. While this could make the generated images adhere to specific logical laws, the diversity is extremely limited, suggesting that this logical relationship cannot be generalized to other contexts or spatial compositions. Other personalization methods use several images to encourage models to learn the corresponding object or style, thus synthesizing related images that preserve high fidelity with original images but in different contexts or with new subjects. However, these personalization methods mainly focus on the features of images, ignoring the underlying logical laws. Furthermore, the great value of the sentences summarizing the common-sense knowledge is entirely missed!

This paper introduces \emph{LogicalDefender}, a simple yet effective approach to enhance the logical generation ability of text-to-image diffusion models. This method leverages human-summarized common-sense knowledge in textual explanations and further strengthens logical understanding with illustrative images. Specifically, we use textual explanations to define a logical embedding representing specific laws, which contains semantic prior extracted from human knowledge. However, using this embedding as instruction can be ambiguous for diffusion models. To address this, we select several images (typically 6-12) that follow logical laws as illustrative images and use the prompt-tuning process to reinforce our logical embedding. During this process, only the logical embedding is trainable, allowing the model to focus on learning logical information. This interaction between textual explanations and illustrative images guides models to comprehend and emphasize logical information. Moreover, to avoid disturbances from object features, such as color and shape, we build a negative guidance path for logical embedding, emphasizing the understanding of logical information. 

Building upon this effective paradigm, the rich common-sense knowledge in texts can be fully exploited with the assistance of consistent illustrative images, just like the mode of knowledge conveyance in classrooms or from ancient times. Furthermore, this learned logical embedding can be attached to enhance the model's logical generation ability during inference.

Experiments demonstrate that our \emph{LogicalDefender} can serve as a general method for improving the model's performance in logical generation with minimal cost and can be effectively applied to other scenarios.

We summarize our main contributions as follows:

\begin{itemize}
	\item We find that adherence to logical laws significantly impacts model performance assessment, as human subconsciousness also judges the logic of an image, similar to features and styles. This reflects current models' ability to understand and follow these deep-rooted regulations. Fortunately, vast common-sense information has been summarized as textual descriptions for easier knowledge conveyance.
 % transmission.
	
	\item We propose a new method, \emph{LogicalDefender}, which aims to learn the specific logical embedding using textual descriptions and illustrative images. A negative guidance path is designed to eliminate the disturbance of unrelated features. This is the first attempt to bridge the logical gap between recent models and human-summarized common-sense knowledge.
	
	\item We conduct extensive experiments on text-based image generations. Results show that our \emph{LogicalDefender} achieves satisfying outcomes and makes great improvement on the logical generating performance. The additional tuning cost is negligible. 
\end{itemize}

\section{Related Work}

\subsection{Text-Guided Synthesis}
Innovative text-to-image synthesis models aim to transform descriptive text into corresponding visual representations \cite{summary,ti1,ti2,ti3,ti4}. This synthesis is primarily explored through two research avenues: direct image generation from text and the refinement of synthesis control via text guidance \cite{survey1,survey2}.

In direct image generation, models like Blended Diffusion \cite{blendeddiffusion} integrate pre-trained DDPM \cite{ddpm} and CLIP \cite{clip} for text-driven image editing. DiffusionCLIP \cite{diffusionclip} enhances image-text alignment, while unCLIP \cite{dalle2} adopts a two-stage generation approach. Imagen surpasses VQ-GAN CLIP \cite{vqganclip}, Latent Diffusion Models \cite{LDM}, and DALLE 2 \cite{dalle2} with its benchmarked diffusion model. Classifier-guided models \cite{classifier-guidance} enhance diffusion model outputs, and classifier-free \cite{classifier-free} methods mix model scores for improved sample quality. GLIDE \cite{glide} and VQ-Diffusion \cite{vqdiffusion} employ guided diffusion and a vector-quantized model, respectively, for realistic image generation from text prompts.
The second path uses pre-trained models for detailed synthesis control. DreamBooth \cite{DreamBooth} and Textual Inversion (TI) \cite{textinversion} use multiple images to modify images of subjects in various scenes and roles. DreamBooth modifies the model structure, while TI optimizes a new word embedding token for each concept. Imagic \cite{imagic} employs text embedding to enable editing of posture and composition within an image.

These advancements indicate significant progress in text-guided image synthesis, with each model contributing to the evolution of the field and the broadening of potential applications.

\subsection{Personalization}
Numerous studies have emerged that utilize text descriptions for image personalization \cite{person1,person2,person3,person4,Parti}. However, personalization can often be an abstract concept that's difficult to accurately convey through text \cite{person7}. The goal of Text-to-Image (T2I) personalization is to learn these abstract concepts from a few sample images and apply these concepts to novel situations \cite{TIGAN,person6}.

Specific models such as TI \cite{textinversion} and DreamBooth have unique approaches to this goal. TI learns textual embeddings that correspond to the pseudo-word and use these embeddings to guide the personalized generation process during inference. On the other hand, DreamBooth \cite{DreamBooth} uses a class noun combined with a unique identifier and further fine-tune the diffusion model to learn the textual information .
Low-Rank Adaptation (LoRA) \cite{lora} uses additional LoRA layers to finetune the base T2I via optimizing only the weight residuals. Elite \cite{elite} focus on using extensive data to build an encoder for concept inversion.

However, these personalization methods mainly focus on the features of images, ignoring the underlying logical laws.

\section{Methodology}
\emph{LogicalDefender} is a new method designed to attach logical information to embeddings and enhance the model's logic-following capability. Given summarized textual descriptions and illustrative images (typically 6-12 images), our goal is to learn a logical embedding that corresponds to specific common-sense knowledge. Our method is applied to Latent Diffusion Models (LDMs) \cite{LDM}, so we first provide some brief background on LDMs (\cref{3.1}), then present our acquisition of initial logical embedding and considerations in prompt design (\cref{3.2}). Finally, we propose a negative-parallel training path in \cref{3.3}, aiming to eliminate the disturbance of unrelated object features and styles. A comparison between logical-classifier and negative-parallel training path is also presented in \cref{3.3}.

\subsection{Latent Diffusion Models}
\label{3.1}
We use Latent Diffusion Models (LDMs) \cite{LDM} as an example to show how our \emph{LogicalDefender} can improve the model's logic-following capability based on a large pretrained diffusion model. LDMs operate in the latent space of pre-trained autoencoders, thus reducing the memory and computing complexity for both training and sampling while retaining quality.

LDMs can be divided into two components, the pre-trained 
%autoencoder
perceptual compression models
(consist of an encoder $\mathcal{E}$ and a decoder $\mathcal{D}$), and a diffusion model.
Firstly, the encoder $\mathcal{E}$ encodes image $x$ into a latent representation $z=\mathcal{E}(x)$. Under this low-dimensional latent space, a diffusion model is trained using a squared error loss to denoise a noised latent code $z_t$ as follows:
\begin{equation}
\mathcal {L}_{L D M}:=\mathbb{E}_{z, \epsilon \sim \mathcal{N}(0,1), t, y}\left[\left\|\epsilon-\epsilon_\theta\left(z_t, t, \phi_\theta(y)\right)\right\|_2^2\right]
\label{eq1}
\end{equation}
where t is the time step, $\epsilon$ is the added noise, 
$\epsilon_\theta$ is noise predicted by the diffusion network, $\phi_\theta(y)$ is a module that maps a conditional input $y$ into a conditional embedding. 
During inference, the noise is iteratively removed from a random noise such finally we arrive at a denoised latent representation, then the decoder $\mathcal{D}$ is utilized to transform this latent code into a image.

In this paper, we conduct our experiment based on the publicly available Stable-Diffusion-v1-5~\cite{LDM} checkpoint, and $y$ is a text prompt.

\subsection{Designing Initialization Tokens and Prompts}
\label{3.2}
Here we aim to optimize our logical embedding by leveraging both the model's prior and the deep information of illustrative images. Instead of fine-tuning all layers, we focus solely on learning our logical embedding to preserve the model's semantic and spatial prior while avoiding ``language drift'' in the diffusion model. However, the best way to utilize the model's prior remains an open question.

\textbf{Initialization tokens.}
Our goal is to extract rich logical information from common-sense knowledge. We use language, the key carrier of observed rules summarized by humans, to describe logical laws. However, these descriptions can be partial, incomplete, and diverse. Therefore, we employ the Large Language Model (LLM), ChatGPT \cite{chatgpt}, to provide brief textual descriptions, such as ``The fruit grows from the branches and is tightly connected to the branches through the fruit stalk.'' We then use a CLIP-based text encoder to transform these descriptions into initial logical embeddings, leveraging the model's semantic prior.
Experiments show that the embeddings initialized from LLM-summarized description have better expressiveness of rules than other partial ones.

\textbf{Designing prompts.}
As shown in \cref{eq1}, the generation can be conditioned by a text prompt $y$. We sample neutral context texts from CLIP \cite{clip} ImageNet templates and adjust them for our task. We represent the logical embedding obtained above as a placeholder string, $\left[ V\right]$.
We initially modify the prompt to ``A photo with the rule of $\left[ V\right]$.'' However, this proves suboptimal. Since we focus on one specific law, we then try injecting related context into the prompt, such as ``Fresh fruits are hanging on the branch with the rule of $\left[ V\right]$.'' However, this approach decreases performance and increases language drift, possibly due to limited tuning data and unfamiliar prompts.
Based on this observation, we design the prompt as ``A photo of fruit on the tree with the rule of $\left[ V\right]$'', adhering to the initial pre-trained form and introducing the scene into the prompts. Notably, this prompt can be divided into two parts: ``A photo of fruit on the tree'' and ``with the rule of $\left[ V\right]$'', which are the contextual and logical prompt separately, denoted as $\mathbf{a}$ and $\mathbf{c}$ in \cref{algo1}.

\subsection{Negative-Parallel Training Path.}
\label{3.3}
From our experience, previous prompt-tuning methods~\cite{DreamBooth, dong2022dreamartist} align the prompt with several images. However, they focus on the object or style, which is generally easier to extract and learn. In contrast, we want to ignore the shallow features as much as possible, prioritizing the underlying logical information instead. Even though our designed initialization tokens and prompts could help with tending to the logical direction, this still decreases performance and appears as a significant issue that deserves attention.

\begin{figure}[tb]
     \centering
     \includegraphics[width=0.9\linewidth]{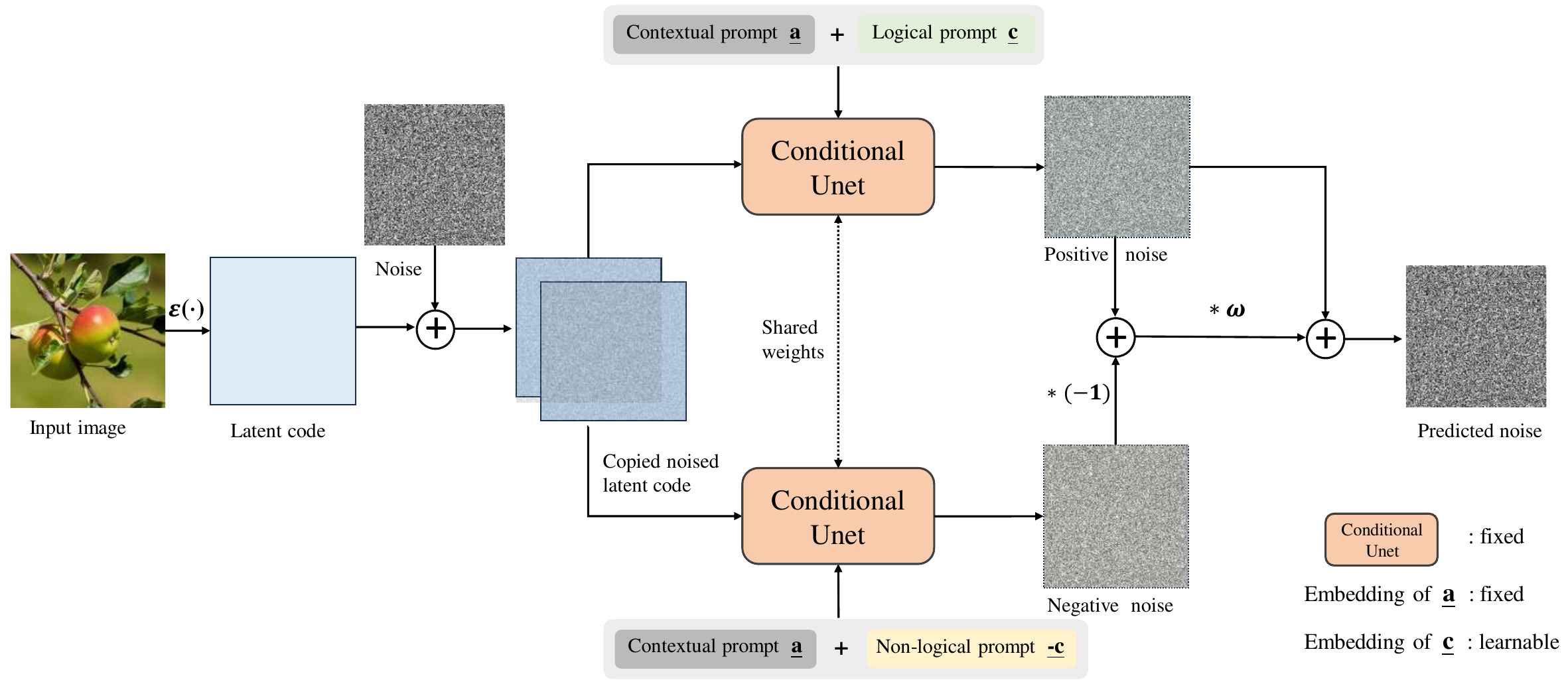}
     \caption{Framework of Negative-Parellel training path. Note that the only trainable parameter in $\mathbf{\theta}$ is our logical embedding, which is under the placeholder string $\left[ V\right]$.}
     \label{fig:negative}
\end{figure}

To tackle this problem, we designed an extra Negative-Parallel training path which helps to remove the shallow features and encourages the model to attend more to the direction of logical laws. 
Specifically, we utilize a non-logical path as the negative one, which depicts the same common-sense knowledge but in the opposite direction.
As shown in \cref{fig:negative}, $\mathbf{a}$ is the prompt that depicts the related contexts and indicates the shallow features, while $\mathbf{c}$ is the logical prompt that conveys the general common-sense knowledge. $\mathbf{-c}$ appears in the Negative-Parallel path, with the prompt: ``not with the rule of $\left[ V\right]$.''.
It would certainly be possible to train separately to get positive and negative embeddings, but we choose joint training because the implementation is extremely simple, and joint-training only one logical embedding tends to interact with both the positive and negative latents together, which condenses the information and does not complicate the training pipeline. Also, we concatenate the positive and negative embeddings together into a single batch, thus only one forward pass is needed, which makes the training pipeline further easier.
Note that the only trainable parameter in $\mathbf{\theta}$ is our logical embedding, which is under the placeholder string $\left[ V\right]$.
And during inference, the negative path is discard.

Our training strategy is shown in \cref{algo1}.
We guide the prompt-tuning process by following a linear combination of the positive and negative estimates:
\begin{equation}
    \Tilde{\mathbf{\epsilon}}_\theta\left(\mathbf{z}_t, \mathbf{a}, \mathbf{c}\right) = \mathbf{\epsilon}_\theta\left(\mathbf{z}_t, \mathbf{a}, \mathbf{c}\right) + \mathbf{\omega}\left[\mathbf{\epsilon}_\theta\left(\mathbf{z}_t, \mathbf{a}, \mathbf{c}\right) - \mathbf{\epsilon}_\theta\left(\mathbf{z}_t, \mathbf{a}, \mathbf{-c}\right)\right]
    \label{eq2}
\end{equation}
Where the positive diffusion score $\mathbf{\epsilon}_\theta(\mathbf{z}_t, \mathbf{a}, \mathbf{c}) \approx -\sigma_t \nabla_{\mathbf{z}_t}\log p(\mathbf{z}_t | \mathbf{a}, \mathbf{c})$ from classifier-free \cite{classifier-free} is modified to move closer in the direction of positive prompts. Thus the latter part can be transformed into:
\begin{equation}
    \mathbf{\omega}\left[\mathbf{\epsilon}_\theta\left(\mathbf{z}_t, \mathbf{a}, \mathbf{c}\right) - \mathbf{\epsilon}_\theta\left(\mathbf{z}_t, \mathbf{a}, \mathbf{-c}\right)\right] \approx -\mathbf{\omega} \sigma_t \nabla_{\mathbf{z}_t}\left[\log p(\mathbf{z}_t | \mathbf{a}, \mathbf{c}) - \log p(\mathbf{z}_t | \mathbf{a}, \mathbf{-c})\right]
    \label{eq3}
\end{equation}

% conditioning prompt set
\begin{algorithm}[tb]
  \caption{Joint training \emph{LogicalDefender} with Negative-Parallel, given a pretrained autoencoder (utilizing an encoder $\mathcal{E}$), diffusion model $\mathbf{f}_\mathbf{\theta}$, and guidance scale $\omega$}
  \label{algo1}
  \begin{algorithmic}[1]
    \Require illustrative data $p(\mathbf{x})$, contextual prompt set $\left\{\mathbf{a}\right\}$, logical prompt $\mathbf{c}$, negative logical prompt $\mathbf{-c}$
    \Repeat
      \State $\mathbf{x} \sim p(\mathbf{x}), \mathbf{a} \sim \left\{\mathbf{a}\right\} $
      \Comment{Sample data and contextual prompt}
      \State $t \sim \rm{Uniform}(\{1, \ldots, T\})$, $\epsilon \sim \mathcal{N}(\mathbf{0}, \mathbf{I})$
      \Comment{Randomly sample time step and noise}
      \State $\mathbf{z}_t \gets \rm{add\ noise}(\mathcal{E}(\mathbf{x}), \mathbf{\epsilon}, t)$ 
      \Comment{Get the noised latent code}
      \State $\mathbf{\epsilon}_\theta\left(\mathbf{z}_t, \mathbf{a}, \mathbf{c}\right) \gets \mathbf{f}_\mathbf{\theta}\left(\mathbf{z}_t, \mathbf{a}, \mathbf{c}\right)$
      \Comment{Obtain noise prediction of the positive path}
      \State $\mathbf{\epsilon}_\theta\left(\mathbf{z}_t, \mathbf{a}, \mathbf{-c}\right) \gets \mathbf{f}_\mathbf{\theta}\left(\mathbf{z}_t, \mathbf{a}, \mathbf{-c}\right)$
      \Comment{Obtain noise prediction of the negative path}
      \State Take gradient step on 
      
      $\grad_\theta \left\| \mathbf{\epsilon}_\theta\left(\mathbf{z}_t, \mathbf{a}, \mathbf{c}\right) + \mathbf{\omega}\left[\mathbf{\epsilon}_\theta\left(\mathbf{z}_t, \mathbf{a}, \mathbf{c}\right) - \mathbf{\epsilon}_\theta\left(\mathbf{z}_t, \mathbf{a}, \mathbf{-c}\right)\right] - \mathbf{\epsilon} \right\|^2$ 
      \Comment{Model optimization}
    \Until{converged}
  \end{algorithmic}
\end{algorithm}

And we find that the contextual prompt $\mathbf{a}$ mainly depicts the related contexts and indicates the shallow features, \eg color, shape \etc , which connects not much with the hidden logical information, conveyed by $\mathbf{c}$. In other words, the logical relation is general, well-suited to various scenarios, and is not closely related to the shallow features, and vice versa. Even if the ``apples'' or ``pears'' exhibit a very normal appearance in color, shape \etc, we still cannot get to the point that this image is logical.
Thus, we hypothesize that this contextual prompt $\mathbf{a}$ is independent of the logical condition $\mathbf{c}$. In this way, we could transform \cref{eq3} into:
\begin{equation}
    -\mathbf{\omega} \sigma_t \nabla_{\mathbf{z}_t}\left[\log \left(\frac{p(\mathbf{z}_t | \mathbf{a}, \mathbf{c})}{p(\mathbf{z}_t | \mathbf{a}, \mathbf{-c})}\right)\right]  = -\mathbf{\omega} \sigma_t \nabla_{\mathbf{z}_t}\left[\log p(\mathbf{z}_t | \mathbf{c}) - \log p(\mathbf{z}_t | \mathbf{-c})\right]
    \label{eq4}
\end{equation}
Thus, the modified diffusion score can be approximated as:
\begin{equation}
    \Tilde{\mathbf{\epsilon}}_\theta\left(\mathbf{z}_t, \mathbf{a}, \mathbf{c}\right) \approx \mathbf{\epsilon}_\theta\left(\mathbf{z}_t, \mathbf{a}, \mathbf{c}\right) + \mathbf{\omega}\left[\mathbf{\epsilon}_\theta\left(\mathbf{z}_t, \mathbf{c}\right) - \mathbf{\epsilon}_\theta\left(\mathbf{z}_t, \mathbf{-c}\right)\right]
    \label{eq5}
\end{equation}
Therefore, it can be more distinctly observed that the predicted diffusion score is guided to move closer in the direction of our logical prompt.
Interestingly, this takes on a role similar to logical classifiers, which could utilize the gradient of the log-likelihood on logical performance to guide the model in generating more logical images.
Even though the extra Negative-Parallel path is different with the trained logical classifier, inspired by classifier-free model \cite{classifier-free}, we could still assume that there was an implicit logical and non-logical classifier separately. And we could get the predicted scores $\mathbf{\epsilon}^*(\mathbf{z}_t, \mathbf{c})$, $\mathbf{\epsilon}^*(\mathbf{z}_t, \mathbf{-c})$, and $\mathbf{\epsilon}^*(\mathbf{z}_t)$. Utilizing the probability formula, 
$p(\mathbf{c} | \mathbf{z}_t) \propto p(\mathbf{z}_t | \mathbf{c})/p(\mathbf{z}_t)$, we could arrive at:
\begin{equation}
    -\sigma_t \nabla_{\mathbf{z}_t}\log p(\mathbf{c} | \mathbf{z}_t ) \approx -\sigma_t \nabla_{\mathbf{z}_t}\left[\log p(\mathbf{z}_t | \mathbf{c}) - p(\mathbf{z}_t)\right]
    \label{eq6}
\end{equation}
Thus:
\begin{equation}
    -\sigma_t \nabla_{\mathbf{z}_t}\log p(\mathbf{c} | \mathbf{z}_t ) \approx -\left[\mathbf{\epsilon}^*(\mathbf{z}_t, \mathbf{c}) - \mathbf{\epsilon}^*(\mathbf{z}_t)\right]
    \label{eq7}
\end{equation}
Similarly, for the non-logical classifier:
\begin{equation}
    -\sigma_t \nabla_{\mathbf{z}_t}\log p(\mathbf{-c} | \mathbf{z}_t ) \approx -\left[\mathbf{\epsilon}^*(\mathbf{z}_t, \mathbf{-c}) - \mathbf{\epsilon}^*(\mathbf{z}_t)\right]
    \label{eq8}
\end{equation}
If we utilize the gradient of the logical and non-logical classifier together, which means, we will use the gradient brought by the logical classifier to move toward the more logical direction. And by using the gradient from a non-logical classifier oppositely, we could also promote more reasonable image generation. The modified diffusion score is:
\begin{equation}
    \Tilde{\mathbf{\epsilon}}_\theta\left(\mathbf{z}_t, \mathbf{a}, \mathbf{c}\right) = \mathbf{\epsilon}_\theta\left(\mathbf{z}_t, \mathbf{a}, \mathbf{c}\right) - \mathbf{\omega} \sigma_t \nabla_{\mathbf{z}_t}\log p(\mathbf{c} | \mathbf{z}_t ) + \mathbf{\omega} \sigma_t \nabla_{\mathbf{z}_t}\log p(\mathbf{-c} | \mathbf{z}_t )
    \label{eq9}
\end{equation}
By substituting \cref{eq7} and \cref{eq8} into the above equation, we get:
\begin{equation}
    \Tilde{\mathbf{\epsilon}}_\theta\left(\mathbf{z}_t, \mathbf{a}, \mathbf{c}\right) \approx \mathbf{\epsilon}_\theta\left(\mathbf{z}_t, \mathbf{a}, \mathbf{c}\right) + \mathbf{\omega}\left[\mathbf{\epsilon}^*\left(\mathbf{z}_t, \mathbf{c}\right) - \mathbf{\epsilon}^*\left(\mathbf{z}_t, \mathbf{-c}\right)\right]
    \label{eq10}
\end{equation}
Remarkably, \cref{eq10} obtained is exactly the same with \cref{eq5}. It is not obvious to find that our Negative-Parallel path displays a similar effect as the pre-trained logical and non-logical classifiers. In comparison, our method is more direct and efficient, saving plenty of computing resources since we do not need to train extra classifiers for assessing logicality based on a large amount of labeled data.

\section{Experiments}
\label{exp}
In this section, we show our experiments and ablation studies. Our method enables the preservation of common-sense knowledge summarized in natural language and can greatly boost the model's logical generation capability through the tuning on illustrative images. We choose to focus on learning the logical relation between fruit and trees.
\subsection{Experimental Settings}

\textbf{Data collection.}
We collected 12 illustrative images for the ``fruit and trees'' logical phenomenon from Pexels~\cite{pexels}, Unsplash~\cite{unsplash} and Pixabay~\cite{pixabay}. It includes 4 kinds of fruits: apples, pears, cherries and lemons. We also designed the contextual prompt set $\{\mathbf{a}\}$. The full list can be found in the Appendix. 

\noindent
\textbf{Implementation details.}
Our experiments are conducted on an A100 with a batch size of 4. The learning rate is 0.0005. We conduct the experiments and ablation studies under 3000 training steps.
The basic guidance scale $\omega$ is set to 3.0 and the number of tokens for logical embedding is set to 6 (see ablations in \cref{4.4})
And we retain the other original settings as LDMs, unless mentioned.

%%%%%%%%%%%%%%%%%%%%
%%% generalization
%%%%%%%%%%%%%%%%%%%%
\begin{figure}[htbp]
\centering
% 第一行
\rotatebox{90}{init} % 行标注
\hspace{0.6mm} % 行标注后的空隙
\begin{minipage}{.22\linewidth}
  \includegraphics[width=\linewidth]{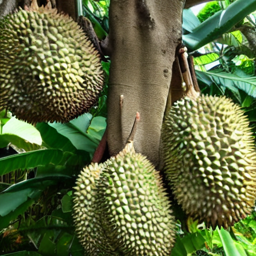}
\end{minipage}\hspace{0.6mm} % 列之间的空隙
\begin{minipage}{.22\linewidth}
  \includegraphics[width=\linewidth]{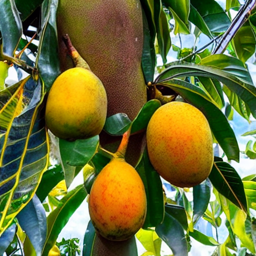}
\end{minipage}\hspace{0.6mm} % 列之间的空隙
\begin{minipage}{.22\linewidth}
  \includegraphics[width=\linewidth]{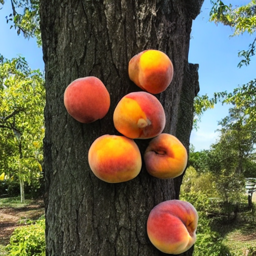}
\end{minipage}\hspace{0.6mm} % 列之间的空隙
\begin{minipage}{.22\linewidth}
  \includegraphics[width=\linewidth]{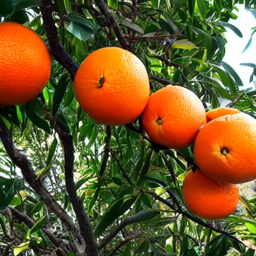}
\end{minipage}
\vspace{0.6mm} % 行之间的空隙

% 第二行
\rotatebox{90}{ours} % 行标注
\hspace{0.6mm} % 行标注后的空隙
\begin{minipage}{.22\linewidth}
  \includegraphics[width=\linewidth]{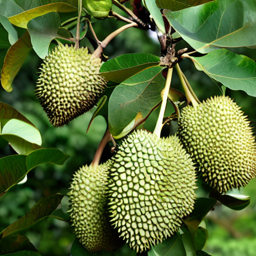}
\end{minipage}\hspace{0.6mm} % 列之间的空隙
\begin{minipage}{.22\linewidth}
  \includegraphics[width=\linewidth]{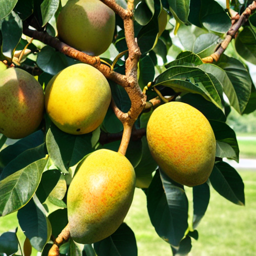}
\end{minipage}\hspace{0.6mm} % 列之间的空隙
\begin{minipage}{.22\linewidth}
  \includegraphics[width=\linewidth]{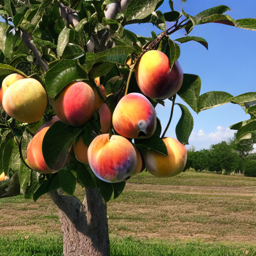}
\end{minipage}\hspace{0.6mm} % 列之间的空隙
\begin{minipage}{.22\linewidth}
  \includegraphics[width=\linewidth]{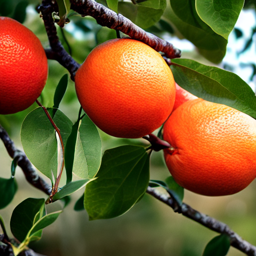}
\end{minipage}

% 列标注
% \newline
\hspace*{1.4cm}durians\hspace{1.5cm} mangos\hspace{1.5cm} peaches \hspace{1.5cm} oranges \hspace{1.4cm}
\caption{Generalization. The figure displays the effective generalization of logical information. Trained on images of apples, cherries, lemons, and pears, our model successfully extrapolates this learning to images of durians, mangoes, peaches, and oranges. All of these fruits as connected to the tree via their stems. However, when inference was made using original SD1.5 model, it resulted in images where fruits appeared to be floating, with no connection to the tree. }
\label{fig:generalization}

\end{figure}

\subsection{Qualitative Results}
Our model, referred to as \emph{LogicalDefender}, was trained on images of apples, cherries, lemons, and pears, and performed inference tasks on these fruit categories. As illustrated in Fig.\ref{fig:init+final}, we compared the logical performance of our model against two other methods: initial results (direct inference from SD1.5) and rule-based results (direct inference from SD1.5 with prompt adjustment). In terms of logical accuracy, our model significantly outperformed the other two methods. For instance, in the case of apples, our model correctly depicted the apple's stem as attached to the tree, unlike the other two methods. Similarly, for cherries, our model accurately represented the cherries as paired and attached to the tree by a branch. For lemons and pears, our model correctly showed the fruits connected to the tree via a stem. These results validate that our model, \emph{LogicalDefender}, significantly enhances the logical coherence of the generated images.

We also discussed the logical generalization of our model . As shown in Fig.\ref{fig:generalization}, the model trained on apples, cherries, lemons, and pears performed well when inferring on durians, mangoes, peaches, and oranges. Direct inference using the SD1.5 model resulted in images where the fruits were not connected to the tree but were instead in a suspended state. However, our model successfully connected these fruits to the tree via stems.

\subsection{Quantitative Results}

\textbf{Evaluation Metrics.}
It is not easy to search for possible quantitative evaluation metrics for logical performance comparing with learning a concept of ``object'' or ``style'', since the logical information is hidden inside and hard to extract by models. To the best of our knowledge, no metrics for measuring the model's logical performance have been proposed. 

However, we do not expect our model ``drifting'' on generating authentic features while learning logical information. 
One important aspect to evaluate is the fidelity between prompt and the image. Thus, we compute the CLIP-T \cite{clip_score} metric, which is the average cosine similarity between prompt and CLIP image embeddings.

\textbf{Results.}
We compare our results with two other methods: initial results and rule-based results of SD1.5 in \cref{tab1}. We can see that our model achieves similar authentic generating capability compared with initial ones, which means that our objective of focus on the logical generation does not affect the normal generation ability of diffusion model.
Other evaluation results and findings on ablation studies can be found in the Appendix.

\begin{table}[tb]
  \caption{Comparison between our model, initial and rule-based results of SD1.5.
  }
  \label{tab1}
  \centering
  \begin{tabular}{lccccccc}
    \toprule
     & \multicolumn{7}{c}{CLIP-T $\uparrow$ }  \\
    Method & apples & pears & cherries &  lemons & oranges  &   apricots & mean \\
    \midrule
    Initial & 29.44 & \textbf{32.32} & \textbf{30.30} & 30.80 & \textbf{30.92} & \textbf{31.23} &  \textbf{30.84} \\
    Rules & 29.25 & 31.54 &  29.65 & 30.18 & 30.38   & 31.13 & 30.36\\
    Ours & \textbf{29.49} & 30.95 &  30.03 & \textbf{30.86} & 30.70   & 30.99 & 30.50\\
    % Heading level & Example & Font size and style\\
    % \midrule
    % 3rd-level heading & {\bf Headings.} Text follows \dots & 10 point, bold\\
    % 4th-level heading & {\it Remark.} Text follows \dots & 10 point, italic\\
  \bottomrule
  \end{tabular}
\end{table}

%%%%%%%%%%%%%%%%%%%%
%%%% 图片个数的讨论
%%%%%%%%%%%%%%%%%%%%
% 左边的两列两行图片
\begin{figure}[htbp]
\centering
% 左边的图片
\begin{minipage}{0.48\textwidth}
  \includegraphics[width=\linewidth]{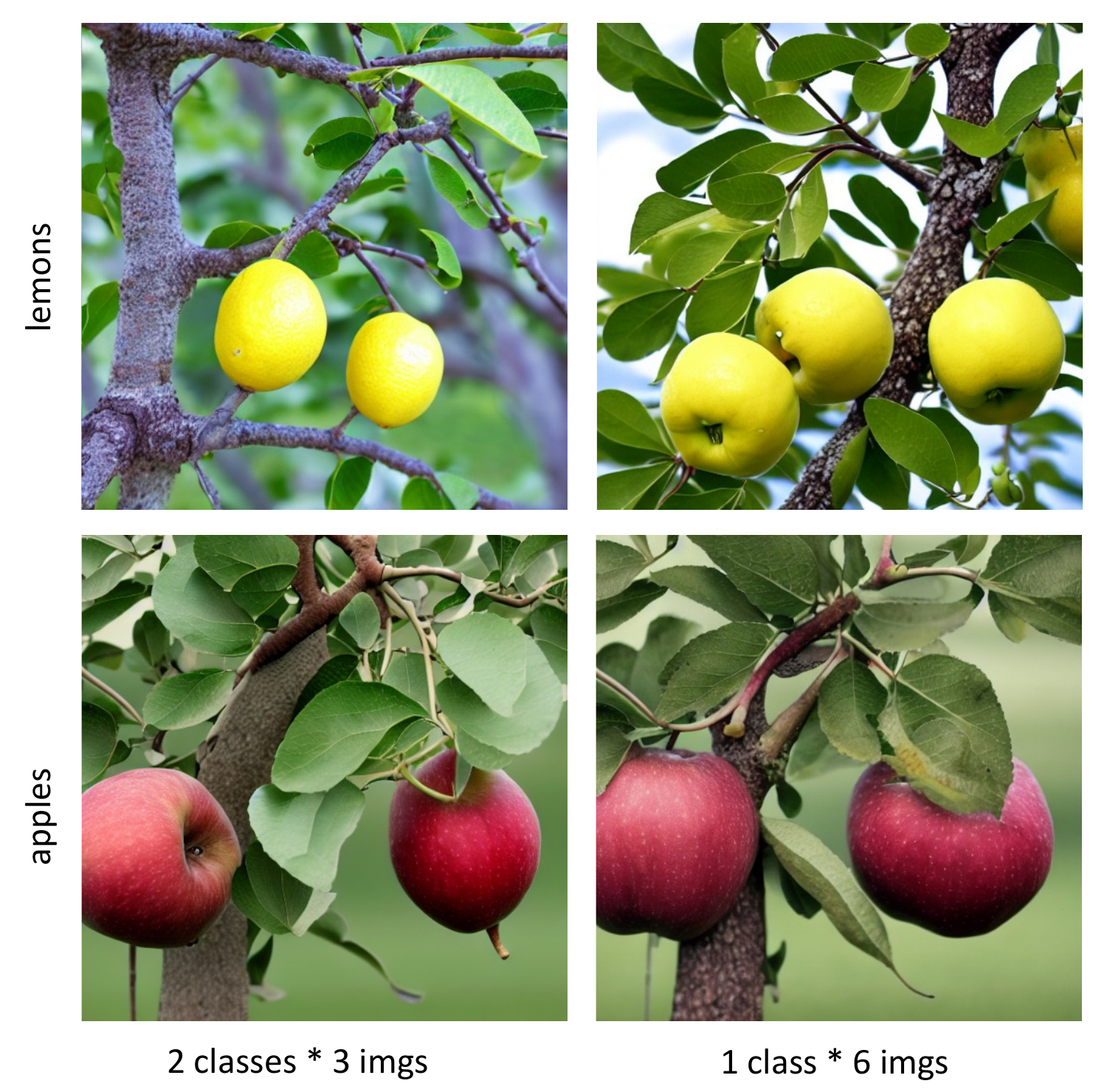}
  \captionof{figure}{Num of classes. ``2 classes * 3 imgs'' represents learning from both lemons and apples, while ``1 class * 6 imgs'' involves learning solely from apple images. The results suggest that increased category diversity improves logical learning and shape preservation. However, while this method enhances apple image outcomes, it lacks effective generalization to other fruits.}
  \label{fig:2_3_1_6}
\end{minipage}\hfill % 使用 \hfill 来添加水平空白，使两个 minipage 分布在页面两侧
% 右边的图片
\begin{minipage}{0.48\textwidth}
  \includegraphics[width=\linewidth]{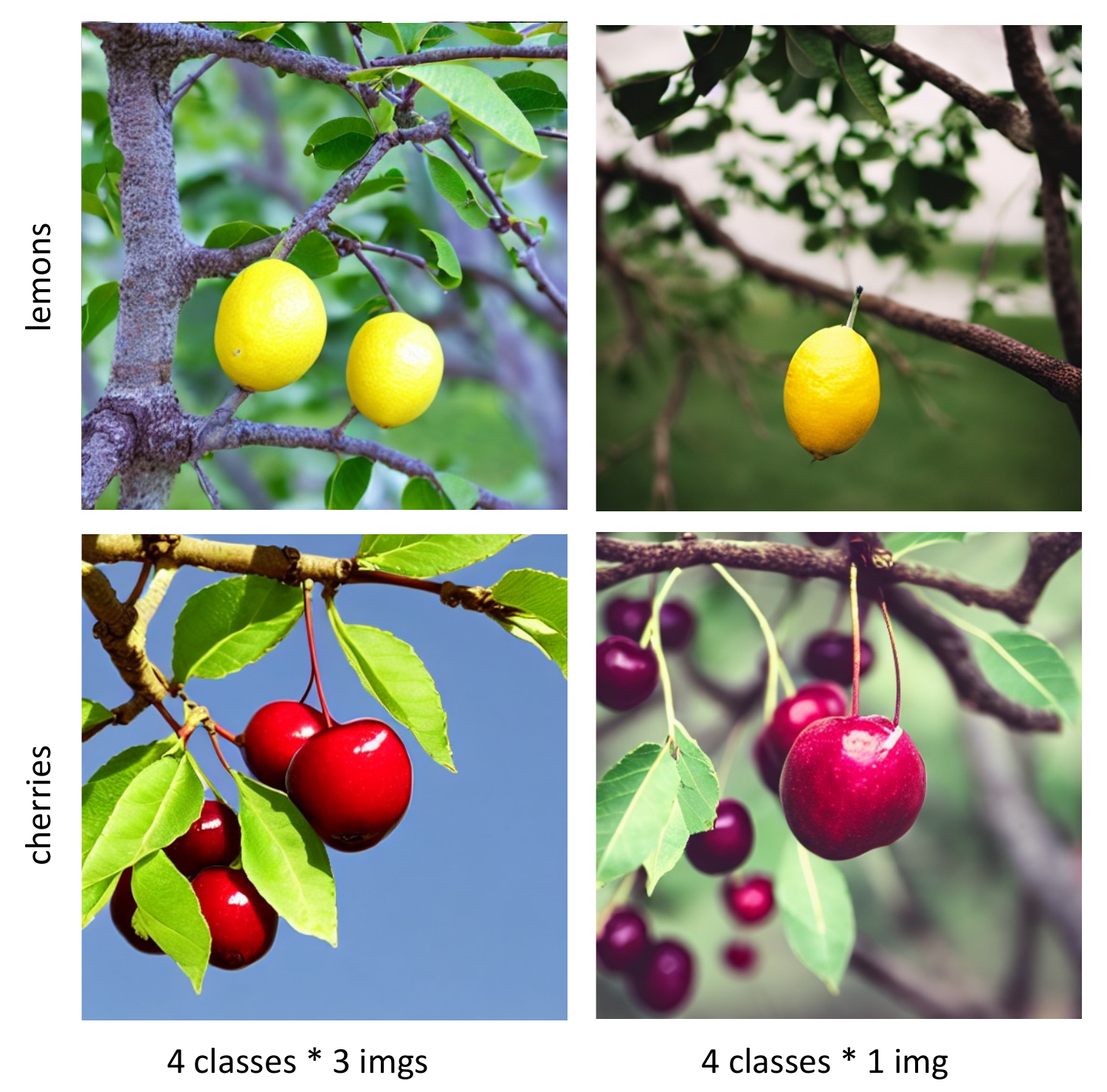}
  \captionof{figure}{Num of images. Learning from three images of each of four fruit types (``4 classes * 3 imgs'') enhances our ability to understand logical information, as evidenced by the learned connection between the object and branches in the lemon and cherry images. This learning outcome is not achieved when only one image per fruit type is used (``4 class * 1 img'').}
  \label{fig:4_3_4_1}
\end{minipage}
\end{figure}

%%%%%%%%%%%%%%%%%%%%
%%%% num token
%%%%%%%%%%%%%%%%%%%%
\begin{figure}[htbp]
\centering
% 第一行
\rotatebox{90}{lemons} % 行标注
\hspace{0.6mm} % 行标注后的空隙
\begin{minipage}{.18\linewidth}
  \includegraphics[width=\linewidth]{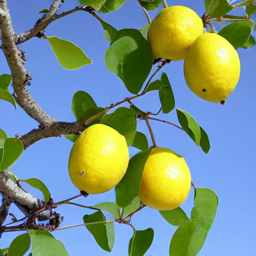}
\end{minipage}\hspace{0.6mm} % 列之间的空隙
\begin{minipage}{.18\linewidth}
  \includegraphics[width=\linewidth]{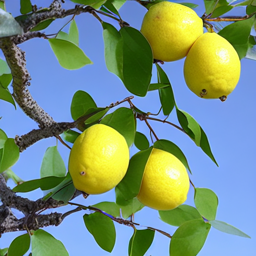}
\end{minipage}\hspace{0.6mm} % 列之间的空隙
\begin{minipage}{.18\linewidth}
  \includegraphics[width=\linewidth]{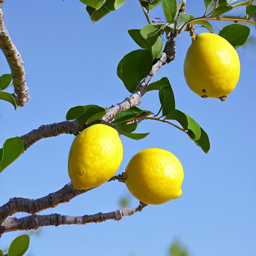}
\end{minipage}\hspace{0.6mm} % 列之间的空隙
\begin{minipage}{.18\linewidth}
  \includegraphics[width=\linewidth]{pic/mean_final_rules_step25/2_lemons.png}
\end{minipage}\hspace{0.6mm} % 列之间的空隙
\begin{minipage}{.18\linewidth}
  \includegraphics[width=\linewidth]{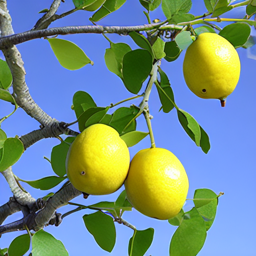}
\end{minipage}
\vspace{0.6mm} % 行之间的空隙

% 第二行
\rotatebox{90}{cherries} % 行标注
\hspace{0.6mm} % 行标注后的空隙
\begin{minipage}{.18\linewidth}
  \includegraphics[width=\linewidth]{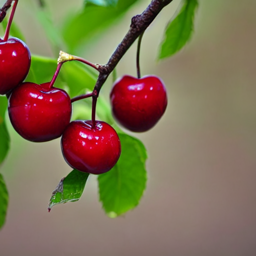}
\end{minipage}\hspace{0.6mm} % 列之间的空隙
\begin{minipage}{.18\linewidth}
  \includegraphics[width=\linewidth]{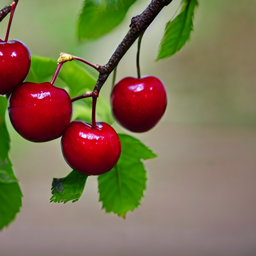}
\end{minipage}\hspace{0.6mm} % 列之间的空隙
\begin{minipage}{.18\linewidth}
  \includegraphics[width=\linewidth]{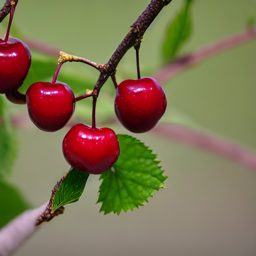}
\end{minipage}\hspace{0.6mm} % 列之间的空隙
\begin{minipage}{.18\linewidth}
  \includegraphics[width=\linewidth]{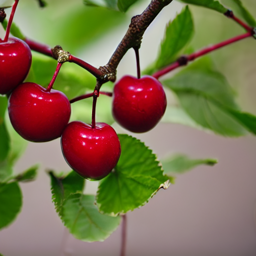}
\end{minipage}\hspace{0.6mm} % 列之间的空隙
\begin{minipage}{.18\linewidth}
  \includegraphics[width=\linewidth]{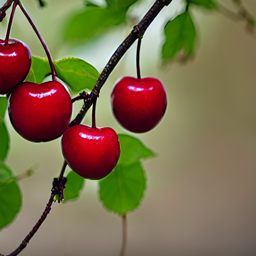}
\end{minipage}

% 列标注
% \newline
\hspace*{0.6cm}1 token\hspace{1.2cm}2 tokens\hspace{1.2cm}3 tokens\hspace{1.2cm}6 tokens\hspace{1.2cm}9 tokens
\caption{Number of tokens. The images display the logical performance of lemons and cherries as the number of tokens increases. As the tokens increase, the lemon image learns to depict the stem as vertically connected to the tree, rather than horizontally or in a zigzag pattern. Similarly, with more tokens, the cherry image more accurately represents the stem as vertically linked to the tree branch, avoiding a zigzag connection.}
\label{fig:tokens}

\end{figure}

\subsection{Ablation Studies}
\label{4.4}

\textbf{Number of tokens.} 
Upon converting the textual description generated by ChatGPT into tokens, we employ a CLIP-based text encoder to transform it into initial logical embedding. Here, we discuss the optimal number of tokens for this embedding. We tested configurations with 1, 2, 3, 6, and 9 tokens. As depicted in Fig.\ref{fig:tokens}, with more tokens, the lemon image accurately depicts the stem as vertically connected to the tree, and the cherry image correctly represents the stem as vertically linked to the tree branch, avoiding unnatural connections.

\begin{figure}[htbp]
\centering
% 左边的图片
\begin{minipage}{0.48\textwidth}
  \includegraphics[width=\linewidth]{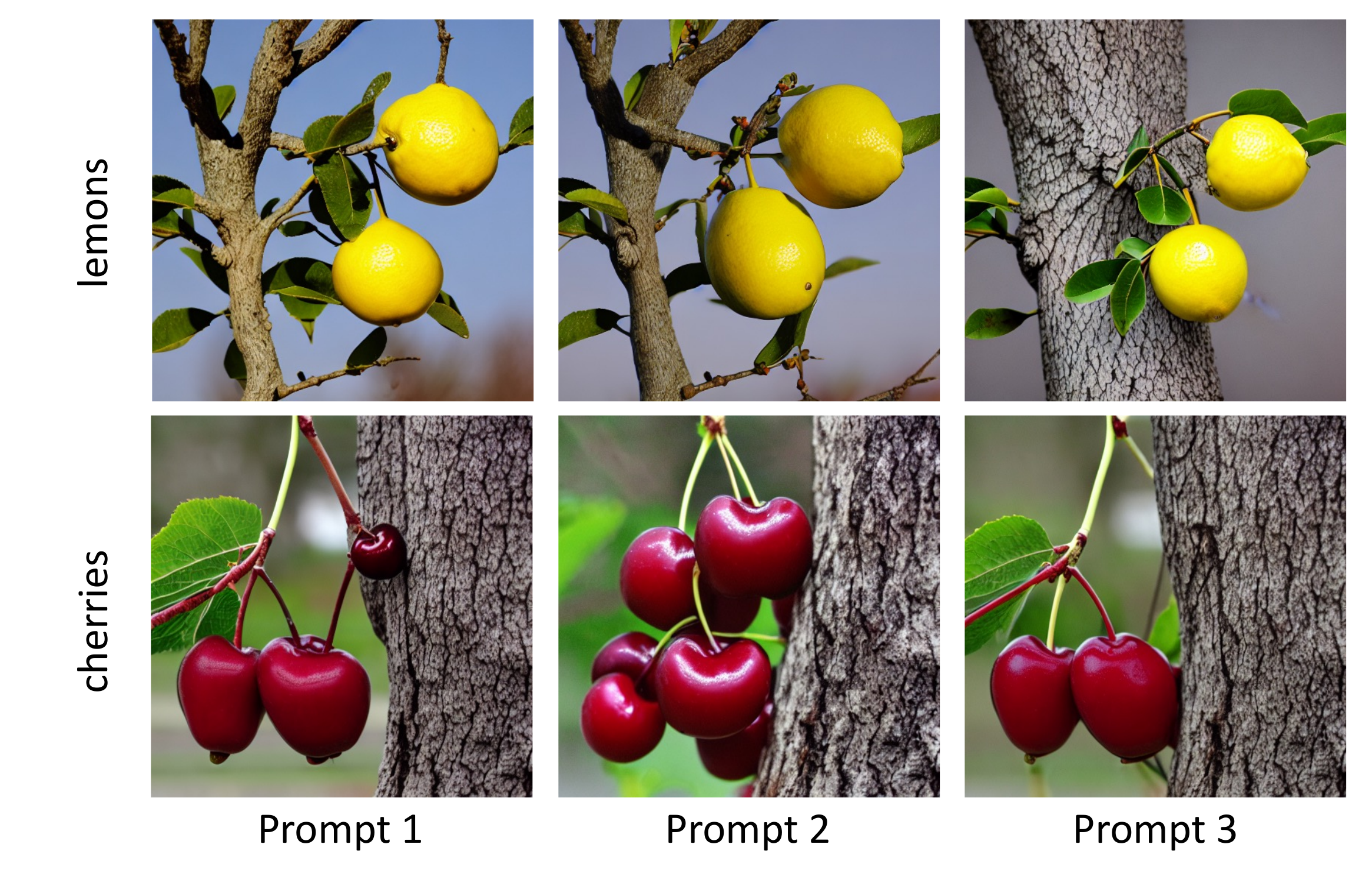}
  \captionof{figure}{Prompt design. The image demonstrates the impact of different prompts on the logical consistency of generated images. Prompt1 is the model's original format, Prompt2 is our custom version with added context, and Prompt3 is our custom prompt that merges the original format with context. In both lemon and cherry images, Prompt3 presents most natural connections between the fruit and tree branches.}
  \label{fig:prompt}
\end{minipage}\hfill % 使用 \hfill 来添加水平空白，使两个 minipage 分布在页面两侧
% 右边的图片
\begin{minipage}{0.48\textwidth}
  \includegraphics[width=0.9\linewidth]{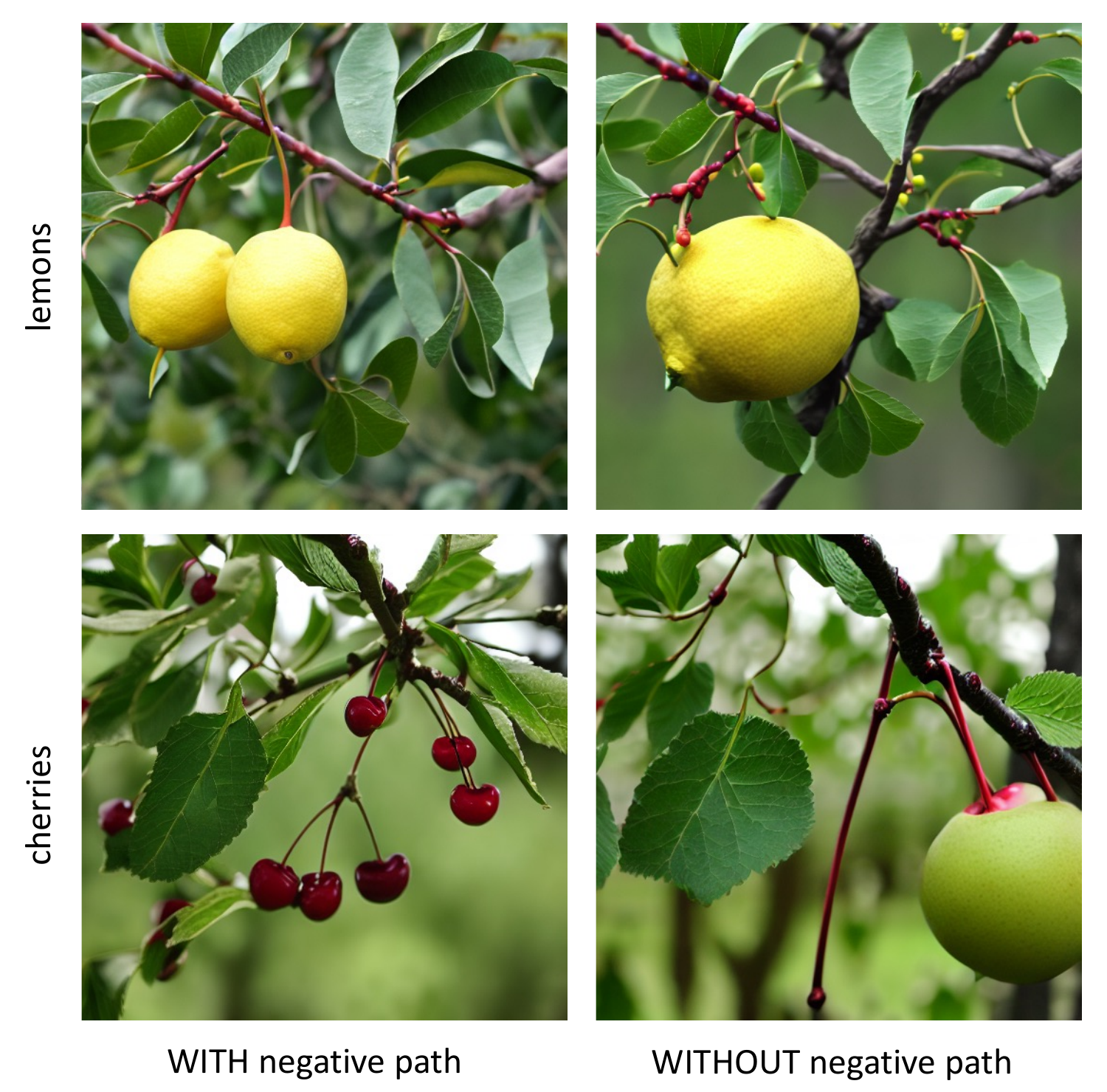}
  \captionof{figure}{Effectiveness of negative-parallel training path on the logical performance of the images. The lemon image demonstrates that the addition of this negative path enables the lemon to learn more accurate logical information without altering its shape. Similarly, the cherry image indicates that incorporating this negative path allows the cherry to acquire improved logical information without changing its color or shape.}
  \label{fig:negative_path}
\end{minipage}
\end{figure}

\noindent
\textbf{Prompt design.}
We investigated prompt designs as depicted in \cref{3.2}. These include the model's inherent prompt1: ``A photo with the rule of $\left[ V\right]$'', our custom prompt2: ``Fresh fruits are hanging on the branch with the rule of $\left[ V\right]$'' infusing with related context for better model guidance, and our custom prompt3: ``A photo of fruit on the tree with the rule of $\left[ V\right]$'' which retains the original pre-trained structure while introducing scenario-related context.
In Fig.\ref{fig:prompt}, we found that prompt3 exhibited superior performance, which depicts a more natural connection between the fruit and tree branches.

%%%%%%%%%%%%%%%%%%%%
%%%% guidance scale
%%%%%%%%%%%%%%%%%%%%

\begin{figure}[htbp]
\centering
% 第一行
\rotatebox{90}{lemons} % 行标注
\hspace{0.4mm} % 行标注后的空隙
\begin{minipage}{.15\linewidth}
  \includegraphics[width=\linewidth]{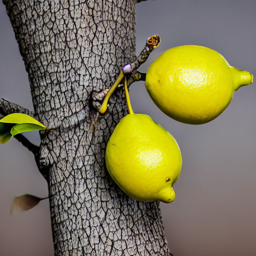}
\end{minipage}\hspace{0.4mm} % 列与列之间的空隙
\begin{minipage}{.15\linewidth}
  \includegraphics[width=\linewidth]{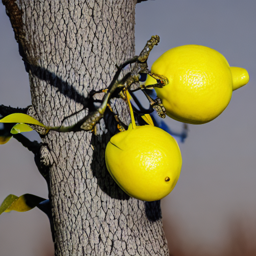}
\end{minipage}\hspace{0.4mm} % 列与列之间的空隙
\begin{minipage}{.15\linewidth}
  \includegraphics[width=\linewidth]{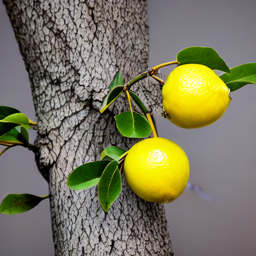}
\end{minipage}\hspace{0.4mm} % 列与列之间的空隙
\begin{minipage}{.15\linewidth}
  \includegraphics[width=\linewidth]{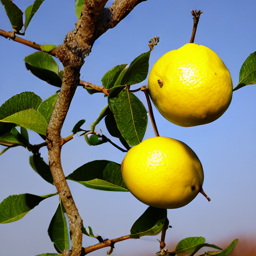}
\end{minipage}\hspace{0.4mm} % 列与列之间的空隙
\begin{minipage}{.15\linewidth}
  \includegraphics[width=\linewidth]{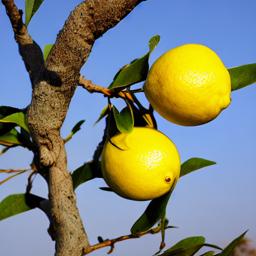}
\end{minipage}\hspace{0.4mm} % 列与列之间的空隙
\begin{minipage}{.15\linewidth}
  \includegraphics[width=\linewidth]{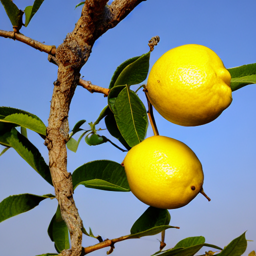}
\end{minipage}
\vspace{0.3mm} % 行与行之间的空隙

% 第二行
\rotatebox{90}{cherries} % 行标注
\hspace{0.4mm} % 行标注后的空隙
\begin{minipage}{.15\linewidth}
  \includegraphics[width=\linewidth]{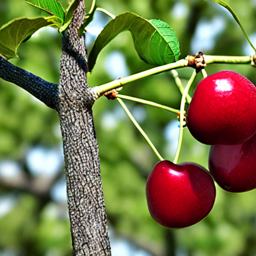}
\end{minipage}\hspace{0.4mm} % 列与列之间的空隙
\begin{minipage}{.15\linewidth}
  \includegraphics[width=\linewidth]{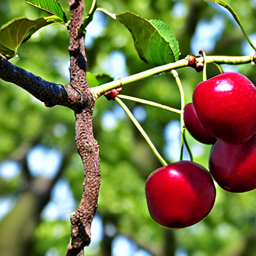}
\end{minipage}\hspace{0.4mm} % 列与列之间的空隙
\begin{minipage}{.15\linewidth}
  \includegraphics[width=\linewidth]{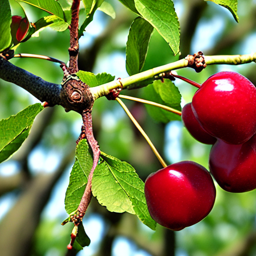}
\end{minipage}\hspace{0.4mm} % 列与列之间的空隙
\begin{minipage}{.15\linewidth}
  \includegraphics[width=\linewidth]{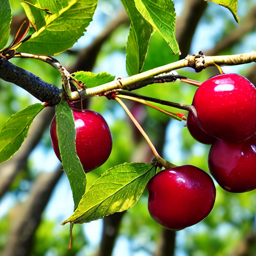}
\end{minipage}\hspace{0.4mm} % 列与列之间的空隙
\begin{minipage}{.15\linewidth}
  \includegraphics[width=\linewidth]{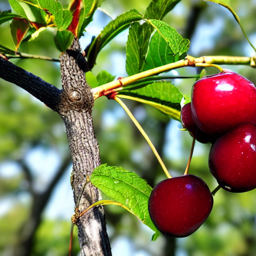}
\end{minipage}\hspace{0.4mm} % 列与列之间的空隙
\begin{minipage}{.15\linewidth}
  \includegraphics[width=\linewidth]{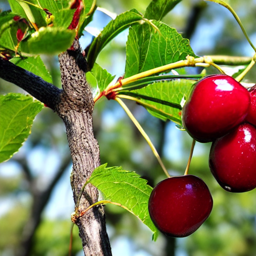}
\end{minipage}

% 列标注
\hspace*{0.9cm}scale=1\hspace{0.9cm}scale=2\hspace{0.9cm}scale=3\hspace{0.9cm}scale=6\hspace{0.9cm}scale=9\hspace{0.9cm}scale=15

\caption{Guidance scale. It displays the logical performance of lemon and cherry images with a guidance scale ranging from 1 to 15. At a scale of 3, the lemon accurately learns the logical information that the tree branches are horizontal, with the stem vertical to the branches, connecting the lemon and the branches. Similarly, the cherry learns the logical information that each cherry corresponds to a single stem. If the scale is too large or too small, the results are not satisfactory.}
\label{fig:guidance_scale}

\end{figure}

%%%%%%%%%%%%%%%%%%%%
%%%% prompt + path
%%%%%%%%%%%%%%%%%%%%
% 左边的两列两行图片

\noindent
\textbf{Number of classes and images.}
Our discussion highlights the impact of image category diversity and the number of images per class during training. As depicted in Fig.\ref{fig:2_3_1_6}, training with a broader variety of image categories enhances logical learning. When trained with both pear and apple images, the model learns the logical connection between the object and branches without distorting the object's shape. However, training with only apple images leads to a distorted lemon representation due to excessive non-logical apple-specific information.
As shown in Fig.\ref{fig:4_3_4_1}, increasing the number of images per class improves the learning of logical information. In the lemon and cherry images, training with a greater number of images helps the model learn the connection between the object and the branches.

% guidance scale / omega  done
\noindent
\textbf{Guidance scale.}
Given that we are using a classifier-free guidance model, we also discuss the impact of guidance size on the logical performance of images.  As in Fig.\ref{fig:guidance_scale}, when the scale is set to 3, the lemon in the first row learns that the tree branches are horizontal, with the stem vertical to the branches, connecting the lemon and the branches. The cherries in the second row, at a scale of 3, learn that each cherry corresponds to a single stem. If the scale is too large or too small, each cherry ends up connected to multiple stems. Therefore, we set the guidance scale to 3.

\noindent
\textbf{Effectiveness of negative-parallel training path.}
We discussed the role of the negative-parallel training path. As shown in Fig.\ref{fig:negative_path}, the performance with the addition of the negative-parallel training path is superior to that without it. The lemon images show that with the addition of the path, the model can better learn the connection between the lemon and the branches. The cherry images show that after adding the path, the model learns the connection between the branches without altering the appearance of the cherries. However, without the negative-parallel training path, the model learns a lot of information beyond logic, such as the shape and color of objects like cherries, lemons, and pears. Thus, adding a negative-parallel training path can effectively suppress the learning of the object's own features, allowing the model to focus solely on logical information.

\section{Discussion}

We would like to emphasize the significance of this ability. It's not just about creating visually appealing images, but about understanding the causal and logical relationships that underpin our world.
Equipping a model with the capability of logical understanding allows it to make sense of the relationships between various elements in the world. This understanding extends beyond the relationships between objects at a given moment, encompassing the relationships between actions, behaviors, and physical world changes across different times. Consequently, the model can plan, reason, and comprehend logic and causality. 

\section{Conclusion}

In this paper, we introduce the importance of logical performance in image generation, as it reflects certain principles of the natural world. We innovatively develop a method, named \emph{LogicalDefender}, to enhance the model's ability to generate images that comply with logical rules. This method strategically designs initial tokens and prompts to learn a logical text embedding that corresponds to specific common-sense knowledge. Furthermore, we introduce a negative guidance path to mitigate the impact of unrelated features in images. Comprehensive experiments focusing on logical performance confirm the effectiveness and generalizability of our method. This highlights the critical importance of logical information and reveals the significant potential of utilizing logical information to generate more realistic images that closely mirror the natural world. More details are available in the Appendix.

% \clearpage\mbox{}Page \thepage\ of the manuscript. This is the last page.
\par\vfill\par
% Now we have reached the maximum length of an ECCV \ECCVyear{} submission (excluding references).
% References should start immediately after the main text, but can continue past p.\ 14 if needed.
% \clearpage  % TODO REVIEW/FINAL: This \clearpage needs to be removed from both review and camera-ready versions.

% ---- Bibliography ----
%
% BibTeX users should specify bibliography style 'splncs04'.
% References will then be sorted and formatted in the correct style.
%
\bibliographystyle{splncs04}
\bibliography{main}

\begin{thebibliography}{10}
\providecommand{\url}[1]{\texttt{#1}}
\providecommand{\urlprefix}{URL }
\providecommand{\doi}[1]{https://doi.org/#1}

\bibitem{pexels}
Pexels. \url{https://www.pexels.com/zh-cn/}

\bibitem{pixabay}
Pixabay. \url{https://pixabay.com/}

\bibitem{unsplash}
Unsplash. \url{https://unsplash.com/}

\bibitem{ti2}
Abdal, R., Zhu, P., Femiani, J.C., Mitra, N.J., Wonka, P.: Clip2stylegan: Unsupervised extraction of stylegan edit directions. ACM SIGGRAPH 2022 Conference Proceedings  (2021), \url{https://api.semanticscholar.org/CorpusID:245117814}

\bibitem{blendeddiffusion}
Avrahami, O., Lischinski, D., Fried, O.: Blended diffusion for text-driven editing of natural images. 2022 IEEE/CVF Conference on Computer Vision and Pattern Recognition (CVPR) pp. 18187--18197 (2021), \url{https://api.semanticscholar.org/CorpusID:244714366}

\bibitem{ti3}
Bar-Tal, O., Ofri-Amar, D., Fridman, R., Kasten, Y., Dekel, T.: Text2live: Text-driven layered image and video editing. ArXiv  \textbf{abs/2204.02491} (2022), \url{https://api.semanticscholar.org/CorpusID:247996703}

\bibitem{survey2}
Cao, H., Tan, C., Gao, Z., Xu, Y., Chen, G., Heng, P.A., Li, S.Z.: A survey on generative diffusion model. ArXiv  \textbf{abs/2209.02646} (2022), \url{https://api.semanticscholar.org/CorpusID:252090040}

\bibitem{person2}
Chang, H., Zhang, H., Barber, J., Maschinot, A., Lezama, J., Jiang, L., Yang, M., Murphy, K.P., Freeman, W.T., Rubinstein, M., Li, Y., Krishnan, D.: Muse: Text-to-image generation via masked generative transformers. ArXiv  \textbf{abs/2301.00704} (2023), \url{https://api.semanticscholar.org/CorpusID:255372955}

\bibitem{person7}
Chen, W., Hu, H., Li, Y., Rui, N., Jia, X., Chang, M.W., Cohen, W.W.: Subject-driven text-to-image generation via apprenticeship learning. ArXiv  \textbf{abs/2304.00186} (2023), \url{https://api.semanticscholar.org/CorpusID:257913352}

\bibitem{person6}
Chen, W., Hu, H., Li, Y., Rui, N., Jia, X., Chang, M.W., Cohen, W.W.: Subject-driven text-to-image generation via apprenticeship learning. ArXiv  \textbf{abs/2304.00186} (2023), \url{https://api.semanticscholar.org/CorpusID:257913352}

\bibitem{vqganclip}
Crowson, K., Biderman, S., Kornis, D., Stander, D., Hallahan, E., Castricato, L., Raff, E.: Vqgan-clip: Open domain image generation and editing with natural language guidance. In: European Conference on Computer Vision (2022), \url{https://api.semanticscholar.org/CorpusID:248239727}

\bibitem{classifier-guidance}
Dhariwal, P., Nichol, A.: Diffusion models beat gans on image synthesis. ArXiv  \textbf{abs/2105.05233} (2021), \url{https://api.semanticscholar.org/CorpusID:234357997}

\bibitem{dong2022dreamartist}
Dong, Z., Wei, P., Lin, L.: Dreamartist: Towards controllable one-shot text-to-image generation via positive-negative prompt-tuning. arXiv preprint arXiv:2211.11337  (2022)

\bibitem{textinversion}
Gal, R., Alaluf, Y., Atzmon, Y., Patashnik, O., Bermano, A.H., Chechik, G., Cohen-Or, D.: An image is worth one word: Personalizing text-to-image generation using textual inversion. ArXiv  \textbf{abs/2208.01618} (2022), \url{https://api.semanticscholar.org/CorpusID:251253049}

\bibitem{vqdiffusion}
Gu, S., Chen, D., Bao, J., Wen, F., Zhang, B., Chen, D., Yuan, L., Guo, B.: Vector quantized diffusion model for text-to-image synthesis. 2022 IEEE/CVF Conference on Computer Vision and Pattern Recognition (CVPR) pp. 10686--10696 (2021), \url{https://api.semanticscholar.org/CorpusID:244714856}

\bibitem{ti4}
Hertz, A., Mokady, R., Tenenbaum, J.M., Aberman, K., Pritch, Y., Cohen-Or, D.: Prompt-to-prompt image editing with cross attention control. ArXiv  \textbf{abs/2208.01626} (2022), \url{https://api.semanticscholar.org/CorpusID:251252882}

\bibitem{clip_score}
Hessel, J., Holtzman, A., Forbes, M., Bras, R.L., Choi, Y.: Clipscore: A reference-free evaluation metric for image captioning. ArXiv  \textbf{abs/2104.08718} (2021), \url{https://api.semanticscholar.org/CorpusID:233296711}

\bibitem{classifier-free}
Ho, J.: Classifier-free diffusion guidance. ArXiv  \textbf{abs/2207.12598} (2022), \url{https://api.semanticscholar.org/CorpusID:249145348}

\bibitem{ddpm}
Ho, J., Jain, A., Abbeel, P.: Denoising diffusion probabilistic models. ArXiv  \textbf{abs/2006.11239} (2020), \url{https://api.semanticscholar.org/CorpusID:219955663}

\bibitem{lora}
Hu, J.E., Shen, Y., Wallis, P., Allen-Zhu, Z., Li, Y., Wang, S., Chen, W.: Lora: Low-rank adaptation of large language models. ArXiv  \textbf{abs/2106.09685} (2021), \url{https://api.semanticscholar.org/CorpusID:235458009}

\bibitem{imagic}
Kawar, B., Zada, S., Lang, O., Tov, O., Chang, H.T., Dekel, T., Mosseri, I., Irani, M.: Imagic: Text-based real image editing with diffusion models. 2023 IEEE/CVF Conference on Computer Vision and Pattern Recognition (CVPR) pp. 6007--6017 (2022), \url{https://api.semanticscholar.org/CorpusID:252918469}

\bibitem{diffusionclip}
Kim, G., Kwon, T., Ye, J.C.: Diffusionclip: Text-guided diffusion models for robust image manipulation. 2022 IEEE/CVF Conference on Computer Vision and Pattern Recognition (CVPR) pp. 2416--2425 (2021), \url{https://api.semanticscholar.org/CorpusID:244909410}

\bibitem{survey1}
Luo, W.: A comprehensive survey on knowledge distillation of diffusion models. ArXiv  \textbf{abs/2304.04262} (2023), \url{https://api.semanticscholar.org/CorpusID:258049177}

\bibitem{glide}
Nichol, A., Dhariwal, P., Ramesh, A., Shyam, P., Mishkin, P., McGrew, B., Sutskever, I., Chen, M.: Glide: Towards photorealistic image generation and editing with text-guided diffusion models. In: International Conference on Machine Learning (2021), \url{https://api.semanticscholar.org/CorpusID:245335086}

\bibitem{chatgpt}
Ouyang, L., Wu, J., Jiang, X., Almeida, D., Wainwright, C.L., Mishkin, P., Zhang, C., Agarwal, S., Slama, K., Ray, A., Schulman, J., Hilton, J., Kelton, F., Miller, L.E., Simens, M., Askell, A., Welinder, P., Christiano, P.F., Leike, J., Lowe, R.J.: Training language models to follow instructions with human feedback. ArXiv  \textbf{abs/2203.02155} (2022), \url{https://api.semanticscholar.org/CorpusID:246426909}

\bibitem{summary}
Radford, A., Kim, J.W., Hallacy, C., Ramesh, A., Goh, G., Agarwal, S., Sastry, G., Askell, A., Mishkin, P., Clark, J., Krueger, G., Sutskever, I.: Learning transferable visual models from natural language supervision. In: International Conference on Machine Learning (2021), \url{https://api.semanticscholar.org/CorpusID:231591445}

\bibitem{clip}
Radford, A., Kim, J.W., Hallacy, C., Ramesh, A., Goh, G., Agarwal, S., Sastry, G., Askell, A., Mishkin, P., Clark, J., Krueger, G., Sutskever, I.: Learning transferable visual models from natural language supervision. In: Proceedings of the International Conference on Machine Learning. vol.~139, pp. 8748--8763 (2021)

\bibitem{dalle2}
Ramesh, A., Dhariwal, P., Nichol, A., Chu, C., Chen, M.: Hierarchical text-conditional image generation with clip latents. ArXiv  \textbf{abs/2204.06125} (2022), \url{https://api.semanticscholar.org/CorpusID:248097655}

\bibitem{person1}
Ramesh, A., Dhariwal, P., Nichol, A., Chu, C., Chen, M.: Hierarchical text-conditional image generation with clip latents. ArXiv  \textbf{abs/2204.06125} (2022), \url{https://api.semanticscholar.org/CorpusID:248097655}

\bibitem{TIGAN}
Reed, S.E., Akata, Z., Yan, X., Logeswaran, L., Schiele, B., Lee, H.: Generative adversarial text to image synthesis. In: Proceedings of the International Conference on Machine Learning. vol.~48, pp. 1060--1069 (2016)

\bibitem{person4}
Richardson, E., Alaluf, Y., Patashnik, O., Nitzan, Y., Azar, Y., Shapiro, S., Cohen-Or, D.: Encoding in style: a stylegan encoder for image-to-image translation. 2021 IEEE/CVF Conference on Computer Vision and Pattern Recognition (CVPR) pp. 2287--2296 (2020), \url{https://api.semanticscholar.org/CorpusID:220936362}

\bibitem{LDM}
Rombach, R., Blattmann, A., Lorenz, D., Esser, P., Ommer, B.: High-resolution image synthesis with latent diffusion models. 2022 IEEE/CVF Conference on Computer Vision and Pattern Recognition (CVPR) pp. 10674--10685 (2021), \url{https://api.semanticscholar.org/CorpusID:245335280}

\bibitem{DreamBooth}
Ruiz, N., Li, Y., Jampani, V., Pritch, Y., Rubinstein, M., Aberman, K.: Dreambooth: Fine tuning text-to-image diffusion models for subject-driven generation. CoRR  \textbf{abs/2208.12242} (2022)

\bibitem{person3}
Tewel, Y., Gal, R., Chechik, G., Atzmon, Y.: Key-locked rank one editing for text-to-image personalization. ACM SIGGRAPH 2023 Conference Proceedings  (2023), \url{https://api.semanticscholar.org/CorpusID:258436985}

\bibitem{elite}
Wei, Y., Zhang, Y., Ji, Z., Bai, J., Zhang, L., Zuo, W.: Elite: Encoding visual concepts into textual embeddings for customized text-to-image generation. 2023 IEEE/CVF International Conference on Computer Vision (ICCV) pp. 15897--15907 (2023), \url{https://api.semanticscholar.org/CorpusID:257219968}

\bibitem{ti1}
Xu, T., Zhang, P., Huang, Q., Zhang, H., Gan, Z., Huang, X., He, X.: Attngan: Fine-grained text to image generation with attentional generative adversarial networks. 2018 IEEE/CVF Conference on Computer Vision and Pattern Recognition pp. 1316--1324 (2017), \url{https://api.semanticscholar.org/CorpusID:8858625}

\bibitem{Parti}
Yu, J., Xu, Y., Koh, J.Y., Luong, T., Baid, G., Wang, Z., Vasudevan, V., Ku, A., Yang, Y., Ayan, B.K., Hutchinson, B., Han, W., Parekh, Z., Li, X., Zhang, H., Baldridge, J., Wu, Y.: Scaling autoregressive models for content-rich text-to-image generation. CoRR  \textbf{abs/2206.10789} (2022)

\bibitem{controlnet}
Zhang, L., Rao, A., Agrawala, M.: Adding conditional control to text-to-image diffusion models. 2023 IEEE/CVF International Conference on Computer Vision (ICCV) pp. 3813--3824 (2023), \url{https://api.semanticscholar.org/CorpusID:256827727}

\end{thebibliography}

\newpage
\renewcommand{\thesubsection}{\Alph{subsection}.} % 修改编号格式为字母

\section*{Appendix}
\subsection{More details for ablation studies}
We performed ablation studies on various factors including prompt design, the number of classes and images, guidance scale, and the effectiveness of the negative-parallel training path. While qualitative results are presented in the main text, we provide quantitative results here.

As discussed in Section 4.3, we use the CLIP-T
metric to measure the fidelity of the object and the prompt. We set the prompt as ``A few \{fruits\} are growing on the tree'' where \{fruits\} is substitute for different categories, and use the CLIP model to infer the image embedding of our generated image. We then calculate the average cosine similarity between the image embedding and the prompt. This cosine similarity, referred to as CLIP-T, is an indicator of how well the image conveys the information in the prompt; a larger value indicates a better match.

In this section, we provide quantitative results of CLIP-T, in terms of the aforementioned factors.

\begin{table}[]
% \begin{table}[tb]
\caption{Quantitative results of CLIP-T metric for ablation studies.}
\label{tab:ablation}
\centering
\renewcommand\arraystretch{1.2}
\setlength{\tabcolsep}{1.25mm}{
\begin{tabular}{lcccccccc}
\toprule
\multicolumn{2}{c}{}                           & \multicolumn{7}{c}{CLIP-T $\uparrow$}                                        \\ \cline{3-9} 
                                 method &  & apples & pears & cherries & lemons & oranges & apricots & means \\ \hline
\multirow{5}{*}{\# of tokens}    & 1  & 29.24  & 32.36 & 30.10    & 31.13  & 30.58   & 30.85    & 30.71 \\
                                  & 2  & 29.60  & 31.95 & 30.06    & 31.41  & 30.78   & 31.19    & 30.83 \\
                                  & 3   & 29.59  & 31.43 & 30.06    & 30.73  & 30.63   & 31.46    & 30.65 \\
                                  & 6  & 29.49  & 30.95 & 30.03    & 30.86  & 30.70   & 30.99    & 30.50 \\
                                  & 9   & 29.30  & 31.19 & 29.99    & 30.60  & 30.45   & 31.37    & 30.48 \\ \hline
\multirow{3}{*}{prompt design}    & p1   & 29.67  & 31.80 & 30.24    & 31.12  & 30.68   & 31.68    & 30.87 \\
                                  & p2   & 29.45  & 30.28 & 30.09    & 30.63  & 30.48   & 31.05    & 30.33 \\
                                  & p3   & 29.49  & 30.95 & 30.03    & 30.86  & 30.70   & 30.99    & 30.50 \\ \hline
\multirow{2}{*}{\# of classes} & 2 * 3  & 29.44  & 30.64 & 29.46    & 29.74  & 30.38   & 30.66    & 30.05 \\
                                  & 1 * 6   & 29.05  & 28.91 & 28.28    & 27.60  & 29.53   & 30.50    & 28.98 \\ \hline
\multirow{2}{*}{\# of images}   & 4 * 3   & 29.49  & 30.95 & 30.03    & 30.86  & 30.70   & 30.99    & 30.50 \\
                                  & 4 * 1   & 28.22  & 28.36 & 28.83    & 28.26  & 30.09   & 29.70    & 28.91 \\ \hline
\multirow{6}{*}{guidance scale}   & 1  & 29.54  & 31.12 & 29.81    & 30.27  & 30.69   & 31.16    & 30.43 \\
                                  & 2  & 29.57  & 30.97 & 29.91    & 30.71  & 30.71   & 30.89    & 30.46 \\
                                  & 3  & 29.49  & 30.95 & 30.03    & 30.86  & 30.70   & 30.99    & 30.50 \\
                                  & 6  & 29.33  & 31.30 & 29.62    & 30.55  & 30.28   & 31.03    & 30.35 \\
                                  & 9  & 29.38  & 31.57 & 30.02    & 30.60  & 30.17   & 30.77    & 30.42 \\
                                  & 15 & 29.04  & 31.27 & 29.86    & 30.18  & 30.04   & 30.86    & 30.21 \\ \hline
\multirow{2}{*}{negative path}    & w/   & 29.49  & 30.95 & 30.03    & 30.86  & 30.70   & 30.99    & 30.50 \\
                                  & w/o  & 29.59  & 31.34 & 29.88    & 30.43  & 30.67   & 31.50    & 30.57 \\ 
\bottomrule                            
\end{tabular}}
\end{table}

Table.\ref{tab:ablation} reveals that the CLIP-T value is unaffected by changes in the initial token design, prompt design, guidance scale, and negative path. This suggests that regardless of variations in these factors, the generated image consistently conveys the information in the prompt accurately.

However, the number of classes used in training impacts the CLIP-T value significantly. Training with 2 classes and 3 images per class results in a larger CLIP-T value than training with 1 class and 6 images. This suggests that training with a single class may introduce additional object-related information, leading to a lower CLIP-T value. In contrast, training with multiple classes allows the model to learn more about the logical knowledge, resulting in a higher CLIP-T value.

Similarly, the number of images used in training also affects the CLIP-T value. Training with 4 classes and 3 images per class yields a higher CLIP-T value than training with 44 classes and 1 image per class. This implies that the more images used in training, the more logical information the model can extract, leading to a higher CLIP-T value.

\subsection{Effect of training steps}
\begin{figure}[htbp]
\centering
% 第一行
% \rotatebox{90}{init} % 行标注
\hspace{0.6mm} % 行标注后的空隙
\begin{minipage}{.22\linewidth}
  \includegraphics[width=\linewidth]{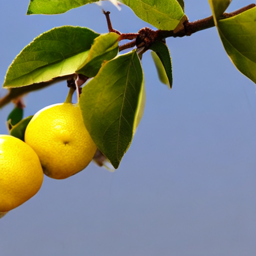}
\end{minipage}\hspace{0.6mm} % 列之间的空隙
\begin{minipage}{.22\linewidth}
  \includegraphics[width=\linewidth]{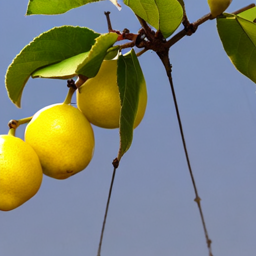}
\end{minipage}\hspace{0.6mm} % 列之间的空隙
\begin{minipage}{.22\linewidth}
  \includegraphics[width=\linewidth]{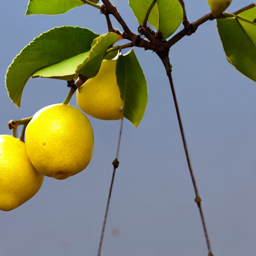}
\end{minipage}\hspace{0.6mm} % 列之间的空隙
\begin{minipage}{.22\linewidth}
  \includegraphics[width=\linewidth]{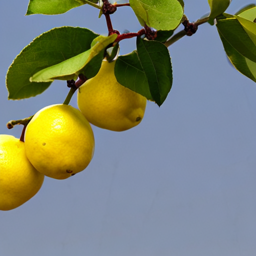}
\end{minipage}
\vspace{1mm} % 行之间的空隙

% \hspace*{0.2cm}500 steps \hspace{1.8cm} 1000\hspace{1.9cm} 1500 \hspace{2cm} 2000 \hspace{2.4cm}
\hspace*{0.7cm}500 steps \qquad \qquad 1000 steps  \qquad \qquad 1500 steps  \qquad \qquad 2000 steps  %\hspace*{0.5cm}
% 第二行
\vspace{1mm} % 行之间的空隙
% \rotatebox{90}{ours} % 行标注
% \hspace{0.6mm} % 行标注后的空隙
\begin{minipage}{.22\linewidth}
\vspace{2.2mm} % 行之间的空隙
  \includegraphics[width=\linewidth]{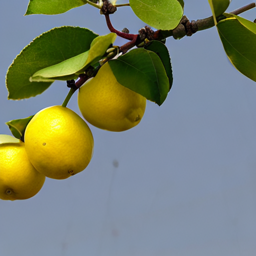}
\end{minipage}\hspace{0.6mm} % 列之间的空隙
\begin{minipage}{.22\linewidth}
\vspace{2.2mm} % 行之间的空隙
  \includegraphics[width=\linewidth]{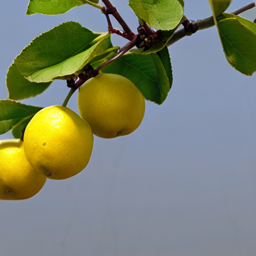}
\end{minipage}\hspace{0.6mm} % 列之间的空隙
\begin{minipage}{.22\linewidth}
\vspace{2.2mm} % 行之间的空隙
  \includegraphics[width=\linewidth]{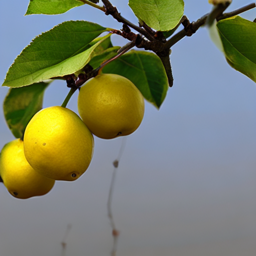}
\end{minipage}\hspace{0.6mm} % 列之间的空隙
\begin{minipage}{.22\linewidth}
\vspace{2.2mm} % 行之间的空隙
  \includegraphics[width=\linewidth]{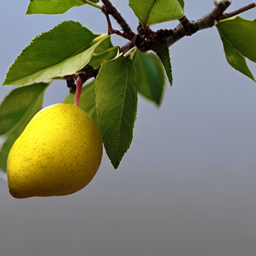}
\end{minipage}

% 列标注
% \newline
\hspace*{0.7cm}3000 steps \qquad \qquad 5000 steps  \qquad \qquad 7000 steps  \qquad \qquad 9000 steps  %\hspace*{1.0cm}
% \hspace*{1.2cm}3000\hspace{1.9cm} 5000\hspace{1.9cm} 7000 \hspace{1.8cm} 9000 \hspace{1.6cm}

\caption{Effect of different training steps on lemons. Notably, after approximately 2000 steps, the model ceases to generate unrealistic features for the lemon tree, such as excessively long branches or angular branches. Therefore, to train this model effectively, around 2000 steps are deemed sufficient.}
\label{fig:training_step_lemon}

\end{figure}

\begin{figure}[htbp]
\centering
% 第一行
% \rotatebox{90}{init} % 行标注
\hspace{0.6mm} % 行标注后的空隙
\begin{minipage}{.22\linewidth}
  \includegraphics[width=\linewidth]{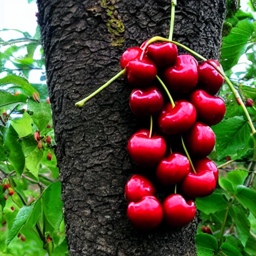}
\end{minipage}\hspace{0.6mm} % 列之间的空隙
\begin{minipage}{.22\linewidth}
  \includegraphics[width=\linewidth]{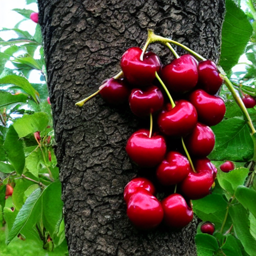}
\end{minipage}\hspace{0.6mm} % 列之间的空隙
\begin{minipage}{.22\linewidth}
  \includegraphics[width=\linewidth]{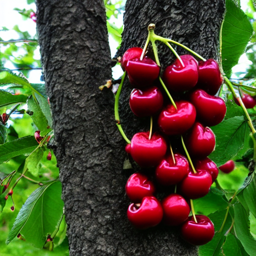}
\end{minipage}\hspace{0.6mm} % 列之间的空隙
\begin{minipage}{.22\linewidth}
  \includegraphics[width=\linewidth]{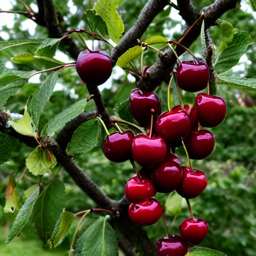}
\end{minipage}
\vspace{1mm} % 行之间的空隙

% \hspace*{0.2cm}500 steps \hspace{1.8cm} 1000\hspace{1.9cm} 1500 \hspace{2cm} 2000 \hspace{2.4cm}
\hspace*{0.7cm}500 steps \qquad \qquad 1000 steps  \qquad \qquad 1500 steps  \qquad \qquad 2000 steps  %\hspace*{0.5cm}
% 第二行
\vspace{1mm} % 行之间的空隙
% \rotatebox{90}{ours} % 行标注
% \hspace{0.6mm} % 行标注后的空隙
\begin{minipage}{.22\linewidth}
\vspace{2.2mm} % 行之间的空隙
  \includegraphics[width=\linewidth]{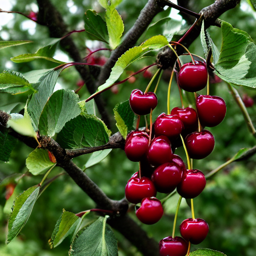}
\end{minipage}\hspace{0.6mm} % 列之间的空隙
\begin{minipage}{.22\linewidth}
\vspace{2.2mm} % 行之间的空隙
  \includegraphics[width=\linewidth]{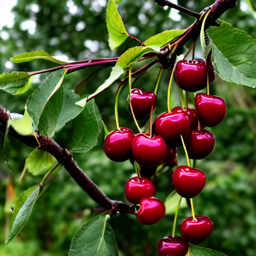}
\end{minipage}\hspace{0.6mm} % 列之间的空隙
\begin{minipage}{.22\linewidth}
\vspace{2.2mm} % 行之间的空隙
  \includegraphics[width=\linewidth]{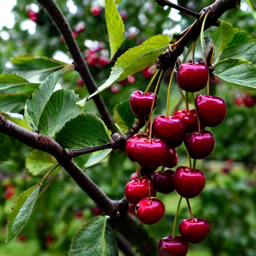}
\end{minipage}\hspace{0.6mm} % 列之间的空隙
\begin{minipage}{.22\linewidth}
\vspace{2.2mm} % 行之间的空隙
  \includegraphics[width=\linewidth]{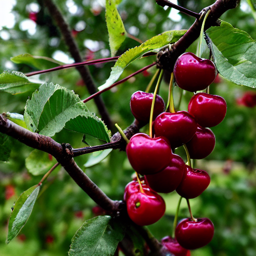}
\end{minipage}

% 列标注
% \newline
\hspace*{0.7cm}3000 steps \qquad \qquad 5000 steps  \qquad \qquad 7000 steps  \qquad \qquad 9000 steps  %\hspace*{1.0cm}
% \hspace*{1.2cm}3000\hspace{1.9cm} 5000\hspace{1.9cm} 7000 \hspace{1.8cm} 9000 \hspace{1.6cm}

\caption{Effect of different training steps on cherries. After approximately 2000 steps, the model learns that cherries should be connected to the tree branches, which in turn are connected to the tree trunk, instead of being directly attached to the tree trunk in a suspended manner. Therefore, around 2000 steps are sufficient to effectively train this model.}
\label{fig:training_step_cherry}
\end{figure}

We also examined the impact of different training steps. Qualitative results of lemons (Fig.\ref{fig:training_step_lemon}) and cherries (Fig.\ref{fig:training_step_cherry}) indicate that the performance stabilizes and differences become negligible after 2000 to 3000 steps. This observation is supported by the quantitative results in Table.\ref{tab:traing_steps}, where the CLIP-T scores are similar across different training steps. This consistency in scores suggests that the generated images effectively convey the information in the prompt, regardless of the number of training steps.

\begin{table}[]
\caption{Comparison of CLIP-T metric for different training steps.}
\label{tab:traing_steps}
\centering
\renewcommand\arraystretch{1.2}
\setlength{\tabcolsep}{1.25mm}{
\begin{tabular}{lcccccccc}
\toprule
\multicolumn{2}{c}{}                         & \multicolumn{7}{c}{CLIP-T $\uparrow$}                                        \\ \cline{3-9} 
\multicolumn{1}{l}{method}      &            & apples & pears & cherries & lemons & oranges & apricots & means \\ \hline
\multirow{9}{*}{training steps} & 500        & 29.10  & 31.82 & 29.93    & 30.80  & 30.38   & 31.03    & 30.51 \\
                                & 1000       & 29.29  & 31.84 & 30.04    & 30.44  & 30.58   & 31.28    & 30.58 \\
                                & 1500       & 29.38  & 31.70 & 29.86    & 30.88  & 30.48   & 31.09    & 30.57 \\
                                & 2000       & 29.37  & 31.66 & 29.79    & 30.77  & 30.61   & 31.03    & 30.54 \\
                                & 3000       & 29.49  & 30.95 & 30.03    & 30.86  & 30.70   & 30.99    & 30.50 \\   
                                & 5000       & 29.60  & 30.91 & 29.88    & 30.56  & 30.71   & 31.48    & 30.52 \\
                                & 7000       & 29.61  & 30.73 & 30.03    & 30.49  & 30.84   & 31.54    & 30.54 \\
                                & 9000       & 29.82  & 31.02 & 29.93    & 30.23  & 30.28   & 31.67    & 30.49 \\
\bottomrule
\end{tabular}}
\end{table}

\subsection{Details about illustrative images}
\begin{figure}[htbp]
\centering
% 第一行
% \rotatebox{90}{1} % 行标注
\begin{minipage}{.21\linewidth}
  \includegraphics[width=\linewidth]{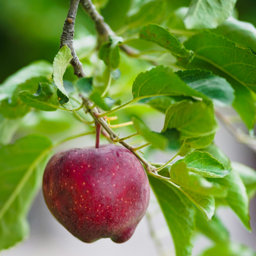}
\end{minipage}%
\hspace{1mm}
\begin{minipage}{.21\linewidth}
  \includegraphics[width=\linewidth]{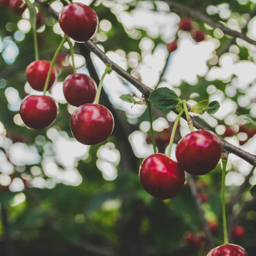}
\end{minipage}%
\hspace{1mm}
\begin{minipage}{.21\linewidth}
  \includegraphics[width=\linewidth]{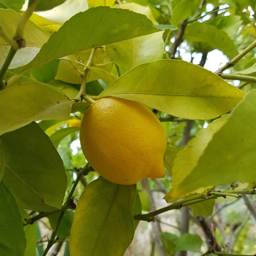}
\end{minipage}%
\hspace{1mm}
\begin{minipage}{.21\linewidth}
  \includegraphics[width=\linewidth]{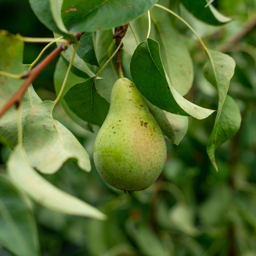}
\end{minipage}
\vspace{1mm}

% 第二行
% \rotatebox{90}{2} % 行标注
\begin{minipage}{.21\linewidth}
  \includegraphics[width=\linewidth]{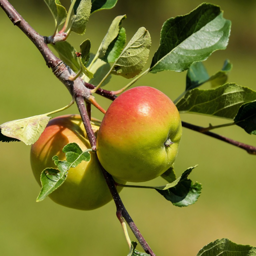}
\end{minipage}%
\hspace{1mm}
\begin{minipage}{.21\linewidth}
  \includegraphics[width=\linewidth]{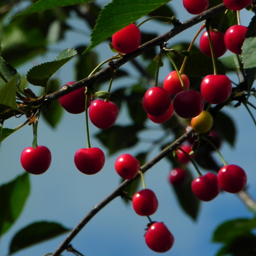}
\end{minipage}%
\hspace{1mm}
\begin{minipage}{.21\linewidth}
  \includegraphics[width=\linewidth]{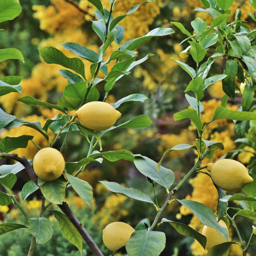}
\end{minipage}%
\hspace{1mm}
\begin{minipage}{.21\linewidth}
  \includegraphics[width=\linewidth]{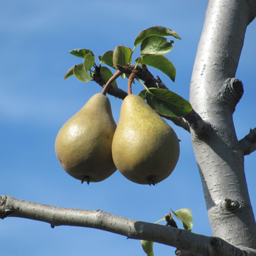}
\end{minipage}
\vspace{1mm}

% 第三行
% \rotatebox{90}{3} % 行标注
\begin{minipage}{.21\linewidth}
  \includegraphics[width=\linewidth]{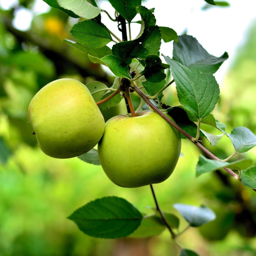}
\end{minipage}%
\hspace{1mm}
\begin{minipage}{.21\linewidth}
  \includegraphics[width=\linewidth]{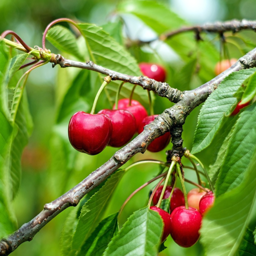}
\end{minipage}%
\hspace{1mm}
\begin{minipage}{.21\linewidth}
  \includegraphics[width=\linewidth]{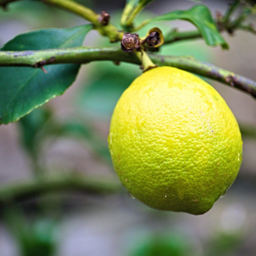}
\end{minipage}%
\hspace{1mm}
\begin{minipage}{.21\linewidth}
  \includegraphics[width=\linewidth]{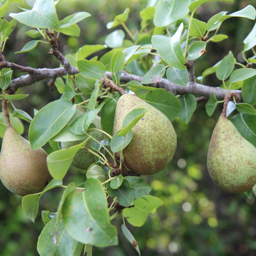}
\end{minipage}

% 列标注
\hspace{1.6cm}apples \hspace{1.6cm}cherries \hspace{1.6cm} lemons\hspace{1.6cm}pears\hspace{1.6cm}
\caption{Our collected real images for logical illustration.}
\label{fig:illustrative_images}
\end{figure}

We collected 12 illustrative images for the ``fruit and trees'' logical phenomenon, including 4 kinds of fruits: apples, pears, cherries and lemons. All the images are from:
Pexels (\url{https://www.pexels.com/zh-cn/}),
Unsplash (\url{https://unsplash.com/}),
Pixabay (\url{https://pixabay.com/}). 
The illustrative images are shown in \cref{fig:illustrative_images}.

\subsection{Contextual prompt set}
% The text prompts of $\mathbf{a}$ are listed below:
The text prompts of $\{\mathbf{a}\}$ are listed below:

\begin{itemize}
    \item ``a photo of fruits on the tree with the rule of $\left[ V\right]$.''
    \item ``a rendering of fruits on the tree with the rule of $\left[ V\right]$.''
    \item ``a cropped photo of fruits on the tree with the rule of $\left[ V\right]$.''
    \item ``the photo of fruits on the tree with the rule of $\left[ V\right]$.''
    \item ``a clean photo of fruits on the tree with the rule of $\left[ V\right]$.''
    \item ``a dirty photo of fruits on the tree with the rule of $\left[ V\right]$.''
    \item ``one photo of fruits on the tree with the rule of $\left[ V\right]$.''
    \item ``a cool photo of fruits on the tree with the rule of $\left[ V\right]$.''
    \item ``a close-up photo of fruits on the tree with the rule of $\left[ V\right]$.''
    \item ``a bright photo of fruits on the tree with the rule of $\left[ V\right]$.''
    \item ``one cropped photo of fruits on the tree with the rule of $\left[ V\right]$.''
    \item ``a good photo of fruits on the tree with the rule of $\left[ V\right]$.''
    \item ``one close-up photo of fruits on the tree with the rule of $\left[ V\right]$.''
    \item ``a rendition of fruits on the tree with the rule of $\left[ V\right]$.''
    \item ``a nice photo of fruits on the tree with the rule of $\left[ V\right]$.''
    \item ``a small photo of fruits on the tree with the rule of $\left[ V\right]$.''
    \item ``a weird photo of fruits on the tree with the rule of $\left[ V\right]$.''
    \item ``a large photo of fruits on the tree with the rule of $\left[ V\right]$.''
\end{itemize}

\end{document}